\documentclass[10pt]{article} 

\usepackage{booktabs} 
\usepackage[table]{xcolor}
\usepackage{xcolor}
\usepackage{comment}
\usepackage{float}

 \usepackage[preprint]{rlc}

\usepackage{amssymb}            
\usepackage{mathtools}          
\usepackage{mathrsfs}           

\usepackage{graphicx}           
\usepackage{subcaption}         
\usepackage[space]{grffile}     
\usepackage{url}                

\usepackage{wrapfig}

\usepackage{amsmath,bm}
\usepackage{amssymb}
\usepackage{mathtools}
\usepackage{amsthm}

\usepackage[capitalize,noabbrev]{cleveref}
\crefname{section}{Sec.}{Secs.}
\Crefname{section}{Section}{Sections}
\Crefname{table}{Table}{Tables}
\crefname{table}{Tab.}{Tabs.}

\newcommand{\V}[1]{{\bm{#1}}}
\newcommand{\Vh}[1]{{\hat{\bm{#1}}}}

\newcommand{\eg}{\textit{e.g.} }
\newcommand{\ie}{\textit{i.e.} }
\definecolor{mygreen}{RGB}{0,128,0}
\definecolor{myred}{RGB}{255,0,0}
\title{Physics-Informed Model and Hybrid Planning for Efficient Dyna-Style Reinforcement Learning}

\author{Zakariae El{\ }Asri  
        \quad\quad\quad\quad\quad
        Olivier Sigaud 
        \quad\quad\quad\quad\quad
        Nicolas Thome\\
    \texttt{\{zakariae.el{\_}asri, olivier.sigaud, nicolas.thome\}@sorbonne-universite.fr} \\
    Sorbonne Universit\'{e}, CNRS, ISIR, F-75005 Paris, France}

\begin{document}
\begin{center}
    \maketitle
\end{center}

\begin{abstract}
Applying reinforcement learning (RL) to real-world applications requires addressing a trade-off between asymptotic performance, sample efficiency, and inference time. 
In this work, we demonstrate how to address this triple challenge by leveraging partial physical knowledge about the system dynamics. Our approach involves learning a physics-informed model to boost sample efficiency and generating imaginary trajectories from this model to learn a model-free policy and Q-function. Furthermore, we propose a hybrid planning strategy, combining the learned policy and Q-function with the learned model to enhance time efficiency in planning. Through practical demonstrations, we illustrate that our method improves the compromise between sample efficiency, time efficiency, and performance over state-of-the-art methods. Code is available at  \url{https://github.com/elasriz/PHIHP/}
\end{abstract}

\section{Introduction}
\label{intro}

Reinforcement learning (RL) has proven successful in sequential decision-making tasks across diverse artificial domains, ranging from games to robotics \citep{Mnih2015HumanlevelCT, Lillicrap2016ContinuousCW, fujimoto2018addressing, haarnoja2018soft}. However, this success has not yet been evident in real-world applications, where RL is facing many challenges \citep{dulac2019challenges}, especially in terms of sample efficiency and inference time needed to reach a satisfactory performance. A limitation of existing research is that most works address these three challenges --~sample efficiency, time efficiency, and performance~-- individually, whereas we posit that addressing them simultaneously can benefit from useful synergies between the leveraged mechanisms.

Concretely, on one side Model-Free Reinforcement Learning (MFRL) techniques excel at learning a wide range of control tasks \citep{Lillicrap2016ContinuousCW, fujimoto2018addressing}, but at a high sample cost. 
On the other side, Model-Based Reinforcement Learning (MBRL) drastically reduces the need for samples by acquiring a representation of the agent-environment interaction \citep{Deisenroth2011PILCOAM, chua2018deep}, but requires heavy planning strategies to reach competitive performance, at the cost of inference time. 

A recent line of works focuses on combining MBRL and MFRL to benefit from the best of both worlds \citep{ha2018world, hafner2019dream, clavera2020model}. Particularly, \cite{byravan2021evaluating, wang2019exploring, Hansen2022tdmpc} combine a learned model and a learned policy in planning, this combination helps improve the asymptotic performance but requires more samples, due to the sample cost of learning a good policy.

\begin{wrapfigure}{r}{0.45\textwidth}
    \centering
    \includegraphics[width=0.45\textwidth]{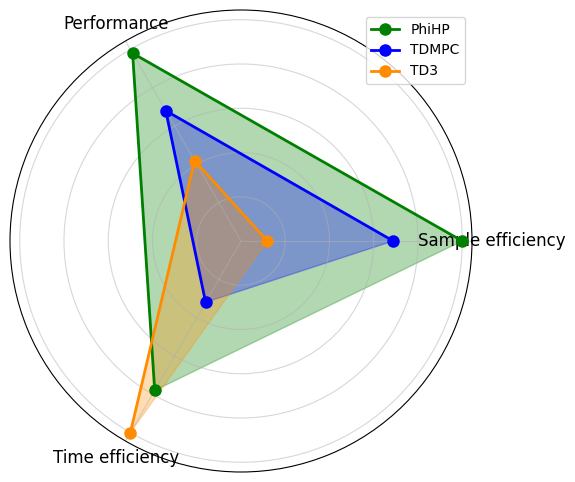}
\caption{PhIHP includes a Physics-Informed model and hybrid planning for efficient policy learning in RL. PhIHP improves the compromise over state-of-the-art methods, model-free TD3 and hybrid TD-MPC, between sample efficiency, time efficiency, and performance. Results averaged over 6 tasks \citep{towers_gymnasium_2023}.}
\label{fig:main}
\vspace{-0.5cm}
\end{wrapfigure}

This paper introduces PhIHP, a {\bf Ph}ysics-{\bf I}nformed model and {\bf H}ybrid {\bf P}lanning method in RL.
PhIHP improves the compromise between the three main challenges outlined above --~sample efficiency, time efficiency, and performance~--, as illustrated in \cref{fig:main}. Compared to state-of-the-art MFRL TD3 \citep{fujimoto2018addressing} and hybrid TD-MPC~\citep{Hansen2022tdmpc}, we show that PhIHP provides a much better sample efficiency, reaches higher asymptotic performance, and is much faster than TD-MPC at inference.

To achieve this goal, PhIHP first learns a physics-informed model of the environment and uses it to learn an MFRL policy in imagination. This policy is used in a hybrid planning scheme. PhIHP leverages three main mechanisms:\\
$\bullet$ {\bf Physics-informed model:} We leverage an approximate physical model and combine it with a learned data-driven residual to match the true dynamics. This physical prior boosts the sample efficiency of PhIHP and the learned residual improves asymptotic performance.\\
$\bullet$ {\bf MFRL in imagination:} we preserve the sample efficiency by training a policy in an actor-critic fashion, using TD3 on trajectories generated from the learned model. The reduced bias in the physics-informed model enables to learn an effective policy in imagination, which is challenging with data-driven models, \eg TD-MPC.\\
$\bullet$ {\bf Hybrid planning strategy:} We incorporate the learned policy and Q-function in planning with the learned model. A better model and policy learned in imagination improve the performance \textit{vs} inference time trade-off.


\section{Related work}
Our work is at the intersection of Model-based RL, physics-informed methods, and hybrid controllers.

{\textbf{Model-based RL:}}
Since  DYNA architectures \citep{sutton1991dyna}, model-based RL algorithms are known to be generally more sample-efficient than model-free methods. Planning with inaccurate or biased models can lead to bad performance due to compounding errors, so many works have focused on developing different methods to learn accurate models: PILCO \citep{Deisenroth2011PILCOAM}, SVG \citep{heess2015learning}, PETS \citep{chua2018deep}, PlaNet \citep{hafner2019learning} and Dreamer \citep{hafner2019dream, hafner2020mastering, hafner2023mastering}. Despite the high asymptotic performance achieved by model-based planning, these methods require a large inference time. 
By contrast, by learning a policy used to sample better actions, we can drastically reduce the inference time.

{\textbf{Physics-informed methods:}}
Recently, a new line of work attempted to leverage the physical knowledge available from the laws of physics governing dynamics, to speed up learning and enhance sample efficiency in MBRL. 
\citep{kloss2017combining, ajay2018augmenting, jeong2019modelling, johannink2019residual, zeng2020tossingbot, cranmer2020lagrangian, yin2021augmenting, yildiz2021continuous, zakariae2022residual, ramesh2023physics}. However, these methods use the learned model in model predictive control (MPC) and suffer from a 
large inference time. 
In this work, we efficiently learn an accurate model by jointly correcting the parameters of a physical prior knowledge and learning a data-driven residual using Neural ODEs.

{\textbf{Hybrid controllers:}}
An interesting line of work consists in combining MBRL and MFRL to benefit from the best of both worlds. This combination can be done 
by using a learned model to generate imaginary samples and augment the training data for a model-free agent \citep{buckman2018sample, clavera2020model, morgan2021model, young2022benefits}. However, the improvement in terms of sample efficiency is limited, since the agent remains trained on real data.
Recent hybrid methods enhance the planning process by using a policy \citep{byravan2021evaluating, wang2019exploring}, or a Q-function \citep{bhardwaj2020blending} with a learned model. More related to our work, TD-MPC \citep{Hansen2022tdmpc} combines the last two methods, using a learned policy and a Q-function with a learned data-driven model to evaluate trajectories. TD-MPC jointly trains all components on real samples and learns a latent representation of the world, resulting in improved sample efficiency. However, the need for samples remains significant as they learn a policy from real data.
By contrast, we first train a physics-informed model from real samples, and then the policy and the Q-function are trained in imagination. In addition, TD-MPC uses an expensive method to optimize sequences of actions, which impacts inference time. By contrast, accurately learning a policy from the physics-informed model reduces the action optimization budget, thereby enhancing time efficiency.

\section{Background}
\label{background}

Our work builds on reinforcement learning and the cross-entropy method.
\paragraph{Reinforcement learning:}
In RL, the problem of solving a given task is formulated as a Markov Decision Process (MDP), that is a tuple $(\mathcal{S}, \mathcal{A}, \mathcal{T}, \mathcal{R}, \gamma, p(s_0))$ where $\mathcal{S}$ is the state space, $\mathcal{A}$ the action space, $\mathcal{T}=: \mathcal{S}\times \mathcal{A} \rightarrow \mathcal{S}$ the transition function, $\mathcal{R}: \mathcal{S}\times \mathcal{A} \rightarrow \mathbb{R}$ the reward function, $\gamma \in [0,1]$ is a discount factor and $\rho_0$ is the initial state distribution.
The objective in RL is to maximize the expected return $\sum_{t=t_0}^{\infty} \gamma^{t-t_0} r_t $ at each timestep $t_0$.
In model-free RL, an agent learns a policy 
$\pi_\theta: \mathcal{S} \rightarrow \mathcal{A}$ that maximizes this expected return.
In contrast, in model-based RL, the agent learns a model that represents the transition function $\mathcal{T}$, then uses this learned model 
$\hat{\mathcal{T}_\theta}$ to predict the next state 
$\hat{s}_{t+1} = \hat{\mathcal{T}_\theta}(s_t,a_t)$. 
The agent maximizes the expected return by optimizing a sequence of actions $A = \{a_{t_0}, ..., a_{t_0+H}\}$ over a horizon $H$:
\begin{equation}
\begin{aligned}
\label{eq:background_optimization}
 ~~~  &A^* =& \mathrm{arg} ~ \underset{A \in {\mathcal{A}^H}}{\max}  ~~ \sum_{t=t_0}^{H} \gamma^{t-t_0} R({s}_t,a_t),  ~~~ \mathrm{subject~to} ~~~  {s}_{t+1} = \hat{\mathcal{T}_\theta}(s_t,a_t).
\end{aligned}
\end{equation}

Furthermore, using an inaccurate model can degrade solutions due to compounding errors. So, one often solves this optimization problem at each time step, only executes the first action from the sequence, and plans again at the next time step with updated state information. This is known as model predictive control (MPC).

\paragraph{Cross Entropy Method (CEM):}
Since the dynamics and the reward functions are generally nonlinear, it is difficult to analytically calculate the exact minimum of \eqref{eq:background_optimization}. 
In this work, we use the derivative-free Cross-Entropy Method \citep{Boer2005ATO} to resolve this optimization problem.
In CEM, the agent looks for the best sequence of actions 
over a finite horizon $H$. It first generates $N$ candidate sequences of actions from a normal distribution $X \sim \mathcal{N}(\mu,\,\sigma^{2})$. 
Then, it evaluates the resulting trajectories using the learned dynamics model using a reward model and determines the $K$ elite sequences of actions $(K < N)$, that is the sequences that lead to the highest return. Finally, the normal distribution parameters $\sigma$ and $\mu$ are updated to fit the elites. This process is repeated for a fixed number of iterations. The optimal action sequence is calculated as the mean of the $K$ elites after the last iteration. We call CEM budget the size of the population times the number of iterations, this budget being the main factor of inference time in methods that use the CEM.

\section{Physics-Informed model for Hybrid Planning}
In this section, we describe PhIHP, our proposed Physics-Informed model for Hybrid Planning. PhIHP first learns a physics-informed residual dynamics model (\cref{sec:phimodel}), then learns a MFRL agent through imagination (\cref{sec:td3imagination}), and uses a hybrid planning strategy at inference (\cref{sec:hybridpolicy}). 
PhIHP follows recent hybrid MBRL/MFRL approaches, \eg TD-MPC \cite{Hansen2022tdmpc}, but the physics-informed model brings important improvements at each stage of the process. It brings a more accurate model, which improves predictive performance and robustness with respect to training data distribution shifts. Crucially, it benefits from the continuous neuralODE method (\cref{sec:phimodel}) to accurately predict trajectories, enabling to learn a powerful model-free agent in imagination (\cref{sec:td3imagination}). Finally, it enables to design a hybrid policy learning (\cref{sec:hybridpolicy}) optimizing the performance \textit{vs} time efficiency trade-off. 
\begin{figure*}[ht]
    \begin{subfigure}{0.33\textwidth}
        \centering
        \includegraphics[width=\linewidth]{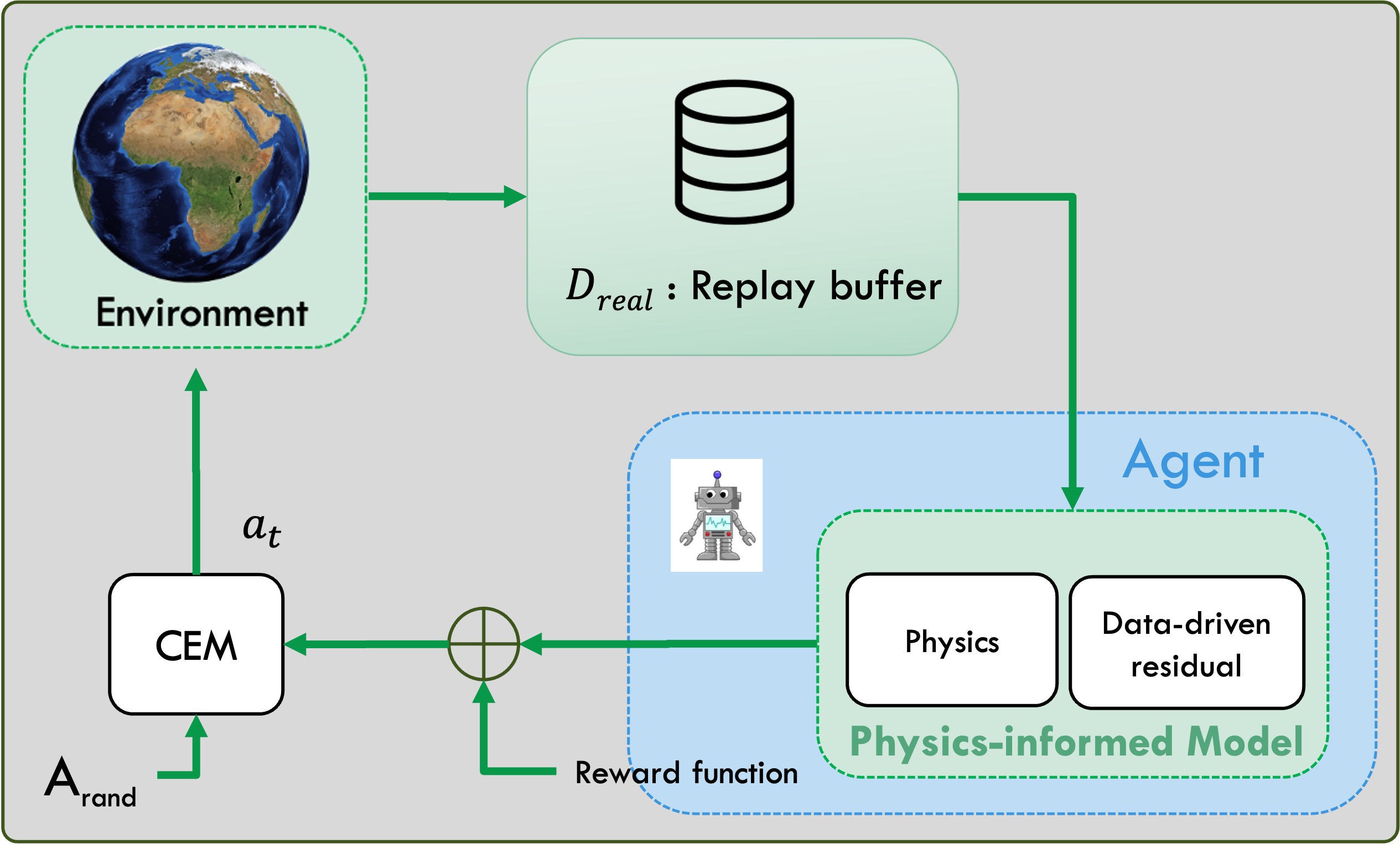}
        \caption{Learn a physics-informed model}
        \label{fig:method_a}
    \end{subfigure}
    \hfill
    \begin{subfigure}{0.33\textwidth}
        \centering
        \includegraphics[width=\linewidth]{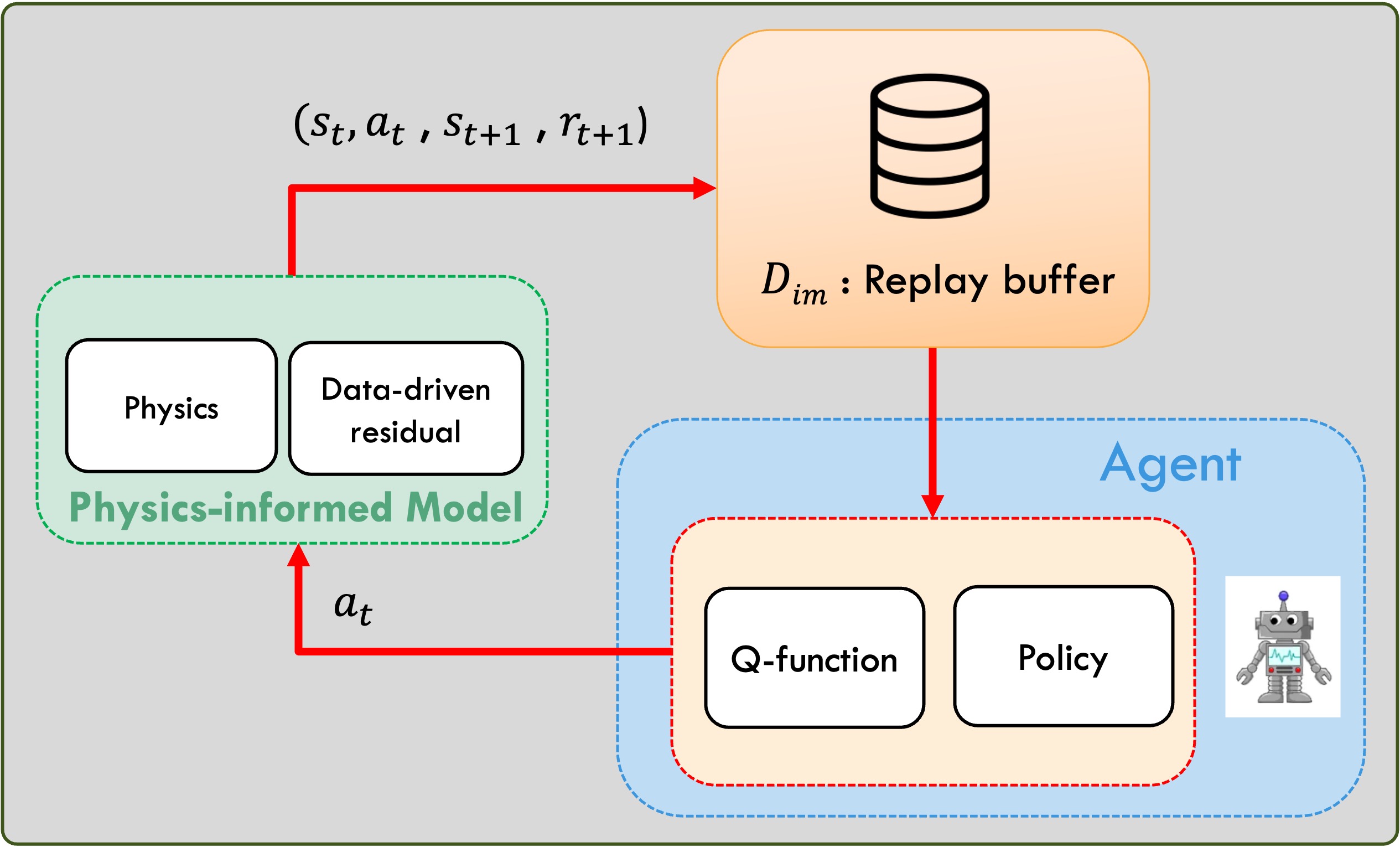}
        \caption{Learn an actor/critic offline}
        \label{fig:method_b}       
    \end{subfigure}
    \hfill
    \begin{subfigure}{0.32\textwidth}
        \centering        \includegraphics[width=\linewidth]{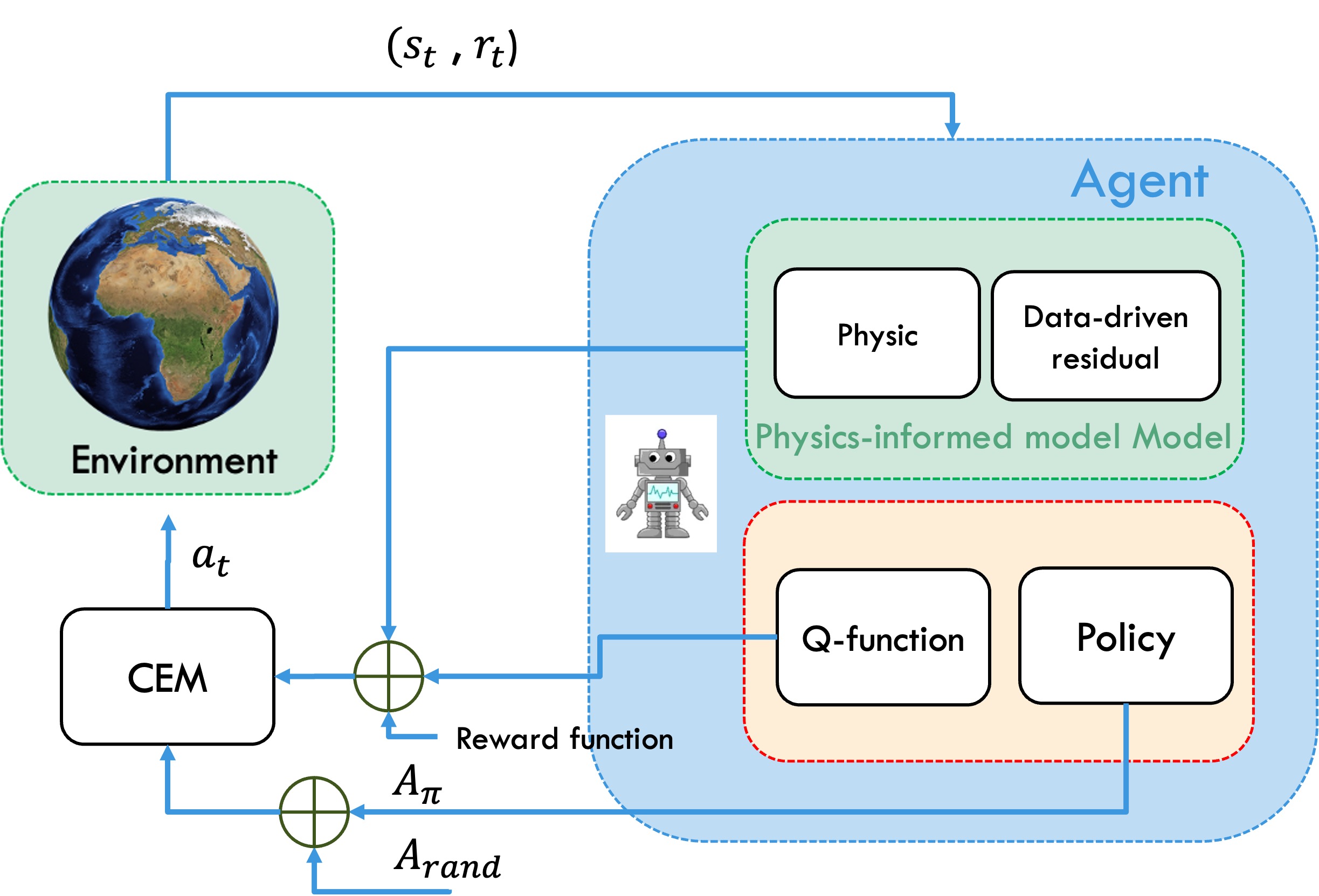}
        \caption{Behaviour at inference time}
        \label{fig:method_c}
    \end{subfigure}
    \caption{\textbf{Schematic view of PhIHP.} (a) We iteratively learn a physics-informed model from few interactions in the environment. (b) We learn a policy and Q-function from trajectories imagined with the learned model. (c) The agent samples actions from the policy output and random actions and then evaluates the resulting trajectories using CEM, a reward function, and the Q-function.}
    \label{fig:method}
\end{figure*}
\subsection{Learning a physics-informed dynamics model}
\label{sec:phimodel}
Model-based RL methods aim to learn the transition function $\mathcal{T}$ of the world \ie a mapping from $(s_t, a_t)$ to $s_{t+1}$. However, learning $\mathcal{T}$ is challenging when $s_{t}$ and $s_{t+1}$ are similar and actions have a low impact on the output, in particular when the time interval between steps decreases.
We address this issue by learning a dynamics function $\hat{\mathcal{T}_\theta}$ to predict the state change $\Delta s_{t}$ over the time step duration $\Delta t$. The next state $s_{t+1}$ can be subsequently determined through integration with an Ordinary Differential Equation (ODE) solver.
Thus, we describe the dynamics as a system following an ODE of the form:
\begin{equation}
\begin{aligned}
 ~~~  &\frac{\mathrm d \V s_t}{\mathrm d t}\Bigr|_{t=t_0} =& \hat{\mathcal{T}_\theta}(\V s_{t_0},\V a_{t_0}),   ~~~ 
\text{and} ~~~  
s_{t+1} \simeq \text{ODESolve}\bigl(\V s_{t},\V a_{t}, \hat{\mathcal{T}_\theta},t,t+\Delta t\bigr),
\label{eq:dynamic}
\end{aligned}
\end{equation}
where $\V s_t$ and $\V a_t$ are the state and action vector for a given time $t$.
We assume the common situation where a partial knowledge of the dynamics is available, generally from the underlying physical laws. The dynamics $\hat{\mathcal{T}_\theta}$ can thus be written as $\hat{\mathcal{T}_\theta} = F_{\theta_p}^p + F_{\theta_r}^r$, where $F_{\theta_p}^p$ is the known analytic approximation of the dynamics and $F_{\theta_r}^r$ is a residual part used to reduce the gap between the model prediction and the real world by learning the complex phenomena that cannot be captured analytically.
The physical model $F_{\theta_p}^p$ is described by an ODE and the residual part $F_{\theta_r}^r$ as a neural network with respective parameters $\theta_p$ and $\theta_r$.
We learn the dynamics model in a supervised manner by optimizing the following objective:
\begin{equation}
\begin{aligned}
\label{eq:lpred}
\mathcal{L}_{pred}(\theta) 
 = \frac{1}{|\mathcal D_{re}|}\sum_{{(s_t,a_t,s_{t+1})} \in {\mathcal D_{re}}} \Vert \V s_{t+1} - \Vh{s}_{t+1} \Vert ^2_2   ~~~ \text{subject to} \quad \frac{\mathrm{d} \Vh s_t}{\mathrm d t}\Bigr|_{t=t'} = (F_{\theta_p}^p + F_{\theta_r}^r)(\V s_{t'},\V a_{t'})\, ,
\end{aligned}
\end{equation}
on a dataset $\mathcal D_{re}$ of real transitions $(s_{t}, a_{t}, s_{t+1})$.
As the decomposition $\hat{\mathcal{T}_\theta} = F_{\theta_p}^p + F_{\theta_r}^r$ is not unique, we apply an $\ell_2$ constraint over the residual part with a coefficient $\lambda$ to enforce the model $\hat{\mathcal{T}_\theta}$ to mostly rely on the physical prior. The learning objective becomes
$\mathcal L_{\lambda}(\theta) = \mathcal{L}_{pred}(\theta) + \frac{1}{\lambda} \cdot \Vert F_{\theta_r}^r \Vert_2$.
The coefficient $\lambda$ is initialized with a value $\lambda_0$ and updated at each epoch with $\lambda_{j+1}=\lambda_{j}+\tau_{ph}\cdot\mathcal{L}_{pred}(\theta)$, where $\lambda_0$ and $\tau_{ph}$ are fixed hyperparameters. 

\subsection{Learning a policy and Q-function through imagination}
\label{sec:td3imagination}

Simply planning with a learned model and CEM is time expensive. 
MFRL methods are generally more time-efficient during inference time than planning methods, since they use policies that directly map a state to an action. However, learning complex policies requires a large amount of training data which impacts sample efficiency. To maintain sample efficiency, a policy can be learned from synthetic data generated by a model. However, an imperfect model may propagate the bias to the learned policy.
In this work, we benefit from the reduced bias in the physics-informed model to generate a sufficiently accurate synthetic dataset $\mathcal D_{im}$ to train a parametric policy $\pi_{\theta}(s_t)$ and a Q-function $Q_{\theta}(s_t,a_t)$, using the TD3 model-free actor-critic algorithm \citep{fujimoto2018addressing}.

\subsection{Hybrid planning with learned model and policy}
\label{sec:hybridpolicy}


PhIHP leverages a hybrid planning method that combines a physics-informed model with a learned policy and Q-function. This combination helps overcome the drawbacks associated with each method when used individually.
While using a sub-optimal policy in control tasks significantly affects the asymptotic performance, planning with a learned model has a high computational cost: $i)$ the planning horizon must be long enough to capture future rewards and $ii)$ the CEM budget 
must be sufficiently large to converge.

We use the learned policy in PhIHP to guide planning. In practice, a CEM-based planner first samples $N_{\pi}$ informative candidates from the learned policy outputs $\hat{\pi}(s_{t})$ and complements them with $N_{rand}$ exploratory candidates sampled from a uniform distribution $X \sim \mathcal{N}(\mu,\,\sigma^{2})$. These informative candidates help reduce the population size and accelerate convergence. The planner estimates the resulting trajectories using the learned model and evaluates each trajectory using the immediate reward function up to the MPC horizon and the Q-value beyond that horizon.

By using the Q-value, we can evaluate the trajectories over a considerably reduced planning horizon $H$ and we add the Q-value of the last state to cover the long-term reward. Hence, the optimization problem is written as follows:
\begin{equation}
\begin{aligned}
\label{eq:optimization}
 &A^* =& \mathrm{arg} ~ \underset{A \in {\mathcal{A}^H}}{\max}   \bigl(~\sum_{t=t_0}^{H} \gamma^{t-t_0} R({s}_t,a_t) +  \alpha\cdot \gamma^{H-t_0} Q(s_H)~\bigr),  ~~~ \mathrm{subject~to} ~~~  {s}_{t+1} = \hat{\mathcal{T}_\theta}(s_t,a_t),
\end{aligned}
\end{equation}
where the discounted sum term represents a local solution to the optimization problem, while the Q-value term encodes the long-term reward and $\alpha$ balances the immediate reward over the planning horizon and the Q-value.

\section{Experiments}

We first compare PhIHP to baselines in terms of performance, sample efficiency, and time efficiency. Then we perform ablations and highlight the generalization capability brought by the physics prior. The robustness of PhIHP to hyper-parameter settings is deferred to Appendix E.

\subsection{Experimental setup}
\label {exp-env}

\textbf{Environments:} We evaluate our method on 6 ODE-governed environments from the gymnasium classic control suite. These include the continuous versions of 3 basic environments: Pendulum, Cartpole, and Acrobot. Additionally, we consider their swing-up variants, where the initial state is “hanging down” and the goal is to swing up and balance the pole at the upright position, similarly to \cite{yildiz2021continuous}. We opted for this benchmark for its challenging characteristics, including tasks with sparse rewards and early termination.

However, to move closer to methods applicable in a real-world situation, we added to the original environments from the gymnasium suite a friction term which is not present in the analytical model of these environments.
Thus, the dynamic of each system is governed by an ODE that can be represented as the combination of two terms: a friction-less component $F^p$ and a friction term $F^r$.
Please refer to Appendix B for additional details.

\textbf{Evaluation metrics.}
In all experiments, we use three main metrics to compare methods:\\
$\bullet$ {Asymptotic performance:} we report the episodic cumulated reward on each environment. \vspace{0.1cm}\\
$\bullet$ {Sample efficiency:} we define the sample efficiency of a method as the minimal amount of samples required to achieve $90\%$ of its maximum performance.\vspace{0.1cm}\\
$\bullet$ {Inference time:} we report the wall-clock time taken by the agent to select an action at one timestep. 

\textbf{Design choice for PhIHP:}
~We learn the model by combining an approximate ODE describing frictionless motion with a data-driven residual model parameterized as a low-dimension MLP.  
We use TD3  \citep{fujimoto2018addressing} for the model-free component of our method, \ie the policy and Q-function. We found it beneficial to modify the original hyperparameters of TD3 to resolve the friction environments. 
For planning, we use CEM-based MPC. Please refer to Appendix C for additional details.

\subsection{Comparison to state of the art:}
\label{sec:sota}
We compare PhIHP to the following state-of-the-art methods: 
\vspace{0.1cm}\\$\bullet$ \textbf{TD-MPC} \citep{Hansen2022tdmpc}, a state-of-the-art hybrid MBRL/MFRL algorithm shown to outperform strong state-based algorithms whether model-based \eg LOOP \citep{sikchi2022learning} and model-free  \eg SAC \citep{haarnoja2018soft} on diverse continuous control tasks. 
\vspace{0.1cm}\\$\bullet$ \textbf{TD3} \citep{fujimoto2018addressing}, a state-of-the-art model-free algorithm. In addition to its popularity and strong performance on continuous control tasks, TD3 is a backbone algorithm for our method to learn the policy and Q-function. We used the same hyperparameters as in PhIHP.\\
\vspace{0.1cm}$\bullet$ \textbf{ CEM-oracle:}  a CEM-based controller with the ground-truth model.

\begin{figure*}[ht]
\centering
    \begin{subfigure}{0.49\textwidth}
        \centering
        \includegraphics[width=\linewidth]{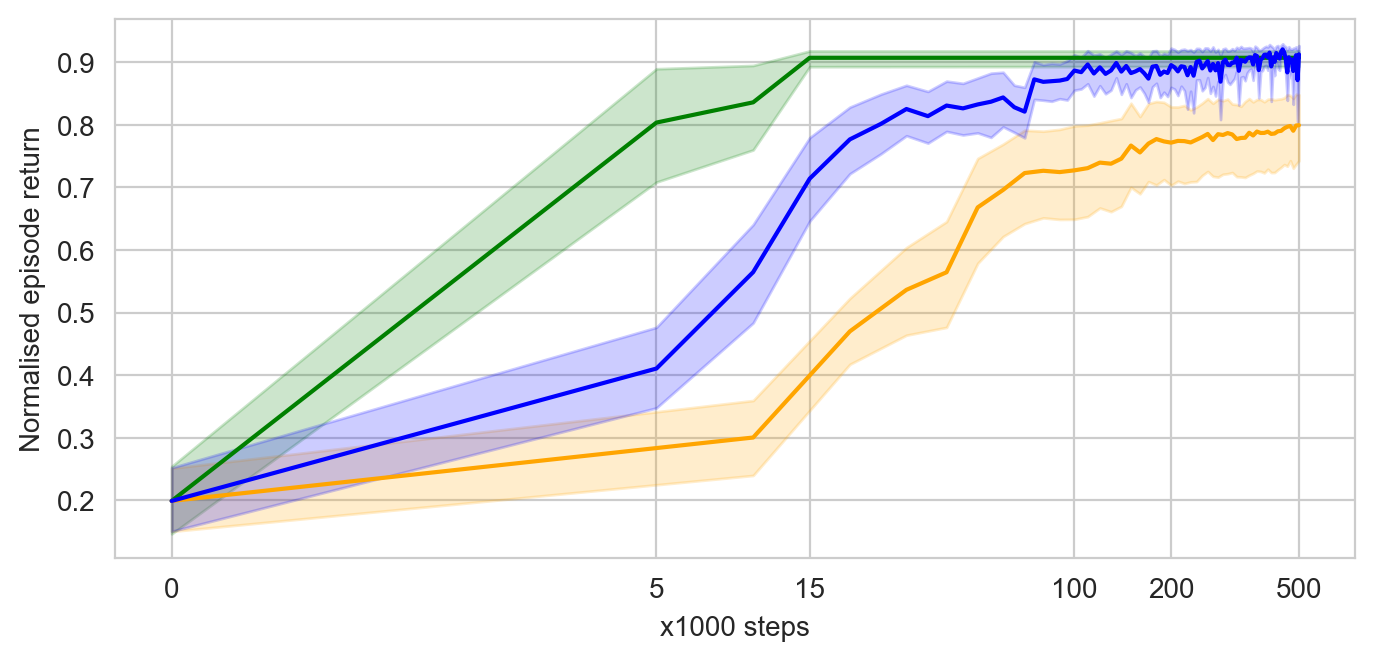}
        \caption{Learning curves, the x-axis uses a symlog scale.}
        \label{figure:learning_curve_scaled}
    \end{subfigure}
    \hfill
    \begin{subfigure}{0.49\textwidth}
        \centering
        \includegraphics[width=\linewidth]{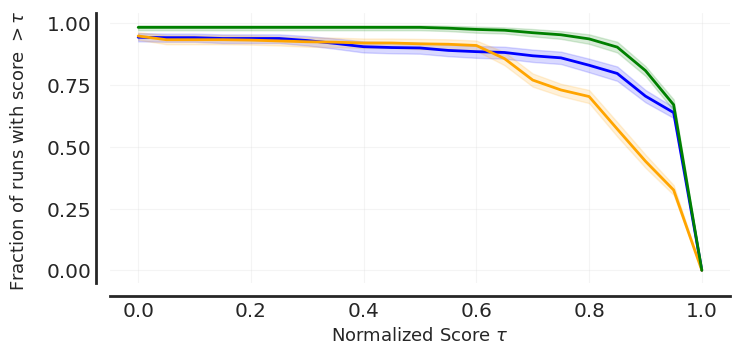}
        \caption{Performance profiles.}
        \label{fig:perf_profiles}        
    \end{subfigure}\\
    \begin{subfigure}{0.4\textwidth}
        \centering
        \includegraphics[width=\linewidth]{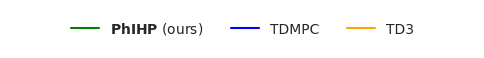}    
    \end{subfigure}
    \caption{Comparison of PhIHP \textit{vs} baselines aggregated on 6 control tasks (10 runs). a) PhIHP shows excellent sample efficiency and better asymptotic performance. b) Performance profiles are obtained with rliable \citep{agarwal2021deep}. PhIHP shows better performance profiles which indicates a better robustness to outliers. Comparison on individual environments are shown in Appendix D
    .}
    \label{fig:test}
\end{figure*}

\begin{table*}[!htbp]

\begin{center}
\begin{small}
\resizebox{\textwidth}{!}
{

\begin{tabular}{|l||cccc||ccc||cccc|}
\toprule


\rowcolor{gray!50}
&\multicolumn{4}{c||}{\textbf{Asymptotic performance}} &\multicolumn{3}{c||}{\textbf{Sample efficiency ($\times 10^3$ samples)}} &\multicolumn{4}{c|}{\textbf{Inference time (in milliseconds)}}\\
\midrule
 & PhIHP& TD-MPC & TD3 & CEM-oracle & PhIHP& TD-MPC & TD3  & PhIHP& TD-MPC & TD3 & CEM-oracle \\
\midrule
Pendulum &  -263 $\pm144$ & -276 $\pm 301$ & -229 $\pm 155$ &-228 $\pm71$
& \textbf{2$\pm0$} & 26$\pm24$ & 86$\pm40$  & 6.32$\pm0.02$ & 39.56$\pm0.28$ & \textbf{0.11}$\pm0.0$ & 18.89$\pm0.26$ \\

Pendulum-sw & \textbf{-356} $\pm 13$&  \textbf{-395$\pm324$} & -368$\pm 14$ & -597 $\pm6$ 
& \textbf{5$\pm0$}  & 28$\pm12$ & 57$\pm16$ & 6.37$\pm0.01$ & 39.6$\pm0.54$ & \textbf{0.11}$\pm0.0$ & 18.87$\pm0.05$ \\

CartPole & \textbf{500$\pm0$} &  432 $\pm129$ & 464$\pm80$  &453$\pm24$ 
& \textbf{5$\pm0$}  & 23$\pm10$ & 108$\pm27$ & 7.43$\pm0.02$ & 39.3$\pm0.07$ & \textbf{0.11}$\pm0.0$ & 33.22$\pm0.03$
\\
CartPole-sw & 453 $\pm 8$ & \textbf{460 $\pm 4 $} & 354 $\pm 113$ &446$\pm5$  
& \textbf{5$\pm0$} & 76$\pm27$  & 27$\pm10$ & 10.13$\pm0.06$ & 39.36$\pm0.05$ & \textbf{0.11}$\pm0.0$ & 33.73$\pm0.03$
\\ 
Acrobot & \textbf{-138 $\pm 122$} & -249$\pm168$ &-237$\pm 183$ &-500$\pm0$
& \textbf{5$\pm0$} & 10$\pm5$  & 233$\pm110$ & 11.14$\pm0.02$ & 39.38$\pm0.09$ & \textbf{0.12}$\pm0.0$ & 59.83$\pm0.1$
\\
Acrobot-sw & \textbf{371 $\pm 52 $}  & \textbf{373 $\pm127$} & 119 $\pm 71 $ &349	$\pm5$
& \textbf{15$\pm0$} & 135$\pm123$ & 500$\pm0$ & 9.12$\pm0.02$ & 39.39$\pm0.06$  & \textbf{0.12}$\pm0.0$ & 58.50$\pm0.27$
\\

\bottomrule
\end{tabular}
}
\end{small}
\end{center}
\caption{Return of PhIHP and baselines on 6 classic control tasks. Mean and std. over 10 runs}
\label{table:sota}
\end{table*}

\begin{figure*}[ht]
\centering
\includegraphics[width=\textwidth]{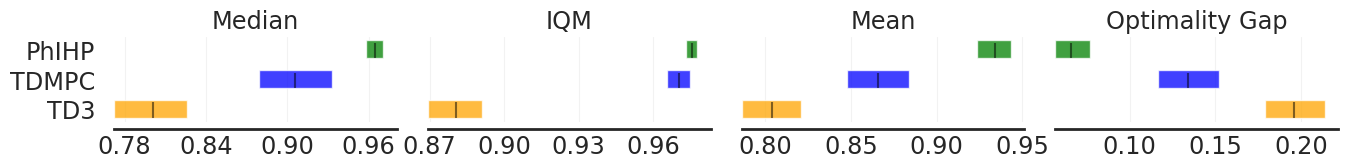}
\caption{Agregated median, interquartile median (IQM), mean performance, and optimality gap of PhIHP and baselines on 6 tasks (10 runs). Higher mean, median, and IQM performance and lower optimality gaps are better. Confidence intervals are estimated using the percentile bootstrap with stratified sampling \citep{agarwal2021deep}. PhIHP outperforms baselines in all metrics.}
\label{figure:rliable_metrics}
\end{figure*}

In \cref{table:sota}, \cref{figure:rliable_metrics} and \cref{figure:learning_curve_scaled}, we show that PhIHP outperforms the baselines with a large margin in at least one of the metrics without being worse on the others. Specifically, PhIHP is far more sample efficient than TD3 and it generally shows 5-15 times better sample efficiency than TD-MPC, except on Acrobot where they are comparable. \cref{figure:learning_curve_scaled} further illustrates this excellent sample efficiency of PhIHP and how TD3 stacks on sub-optimal performance. This enhanced sample efficiency of PhIHP results from training the model-free policy on imaginary trajectories generated by the learned model, as opposed to using real samples in the baselines.
Besides, PhIHP demonstrates superior performance in sparse-reward early-termination environment tasks (Cartpole and Acrobot) compared to TD-MPC, and PhIHP outperforms TD3 with a large margin in Cartpole-swingup, Acrobot, and Acrobot-swingup. Figure 4 in Appendix D.1 shows how TD3 stacks on lower asymptotic performance for the aforementioned tasks. It also shows that TD-MPC performance drops in sparse-reward early-termination environments \eg Cartpole and Acrobot. It also illustrates that, since CEM-oracle uses the reward function to evaluate trajectories within a limited horizon, it manages to solve both tasks with smooth reward functions, and tasks with sparse reward where the goal is to maintain an initial state (\ie Cartpole), but it fails to solve sparse reward problems where the goal is to reach a position out of the planning horizon (\ie Acrobot).

Finally, \cref{fig:perf_profiles} shows that PhIHP has better performance profiles compared to baselines which indicates better robustness to outliers in PhIHP.

\cref{table:sota} also reports the time needed for planning at each time step, obtained with an Apple M1 CPU with 8 cores. It is noteworthy that PhIHP significantly reduces the inference time when compared to TD-MPC. The inference time is still larger than that of TD3 since the latter is a component of our method, but it meets the real-time requirements of various robotics applications. 

\subsection{Ablation study}
In this section, we study the impact of each PhIHP component to illustrate the benefits of using an analytical physics model, imagination learning, and combining CEM with a model-free policy and Q-function for planning. To illustrate this, we compare PhIHP to several methods:

$\bullet$ {\textbf{TD-MPC*:}} our method without physical prior and without imagination. It is similar to TD-MPC since the model is data-driven and it is learned with the policy from real trajectories. But learning the model and the policy are separated.\\
$\bullet$ {\textbf{Ph-TD-MPC*:}} our method without learning in imagination, thus a physics-informed TD-MPC*.\\
$\bullet$ {\textbf{dd-CEM:}} our method without physical prior nor policy component, thus a CEM with a data-driven model learned from real trajectories.\\
$\bullet$ {\textbf{Ph-CEM:}} our method without the policy component, thus a simple CEM with a physics-informed model learned from real trajectories.

\begin{figure*}[ht]
\centering
\includegraphics[width=\textwidth]{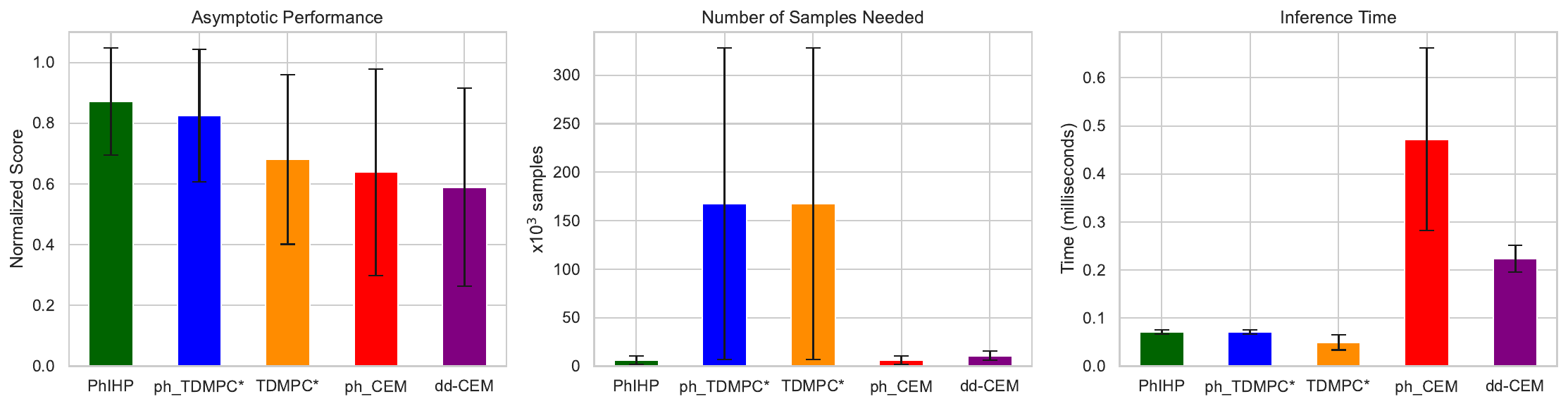}
\caption{Comparison of PhIHP and its variants on the 3 main metrics. The figures illustrate the aggregated results of running all algorithms on 6 classic control tasks. Histograms and bars represent mean and std. over 10 runs.}
\label{fig:ablation}
\end{figure*}

\cref{fig:ablation} shows the impact of the quality of the model on the final performance in MBRL. Precisely, leveraging a physical prior in Ph-CEM and Ph-TD-MPC* shows improvements compared to full data-driven methods, i.e. dd-CEM and TD-MPC*.
We also illustrate that planning with a model, a Q-function, and a policy leads to better performance compared to planning only with the model. For instance Ph-TD-MPC* outperforms Ph-CEM and TD-MPC* outperforms dd-CEM. However, this gain in performance comes with a significant cost in samples, because the agent needs a large amount of data to learn a good policy and Q-function.\\
\cref{fig:ablation} illustrates the trade-off between asymptotic performance, sample efficiency, and inference time in RL. On one hand, methods that learn a model and directly plan with it (\eg dd-CEM and ph-CEM) do not need many samples to achieve sufficiently good performance, but they are too expensive at inference time. On the other hand, methods that learn to plan with a model, Q-function, and policy plan fast but require many samples to train their policies and Q-functions. PhIHP is the only method that achieves good asymptotic performance with low cost in sample efficiency due to learning in imagination and a good inference time due to hybrid planning.


\subsection{Generalization benefits of the physics prior}

In this section, we highlight the key role of incorporating physical knowledge into PhIHP in finding the better compromise between asymptotic performance, sample efficiency, and time efficiency illustrated in \cref{fig:ablation}.
\begin{figure}[h]
\centering
\includegraphics[width=\columnwidth]{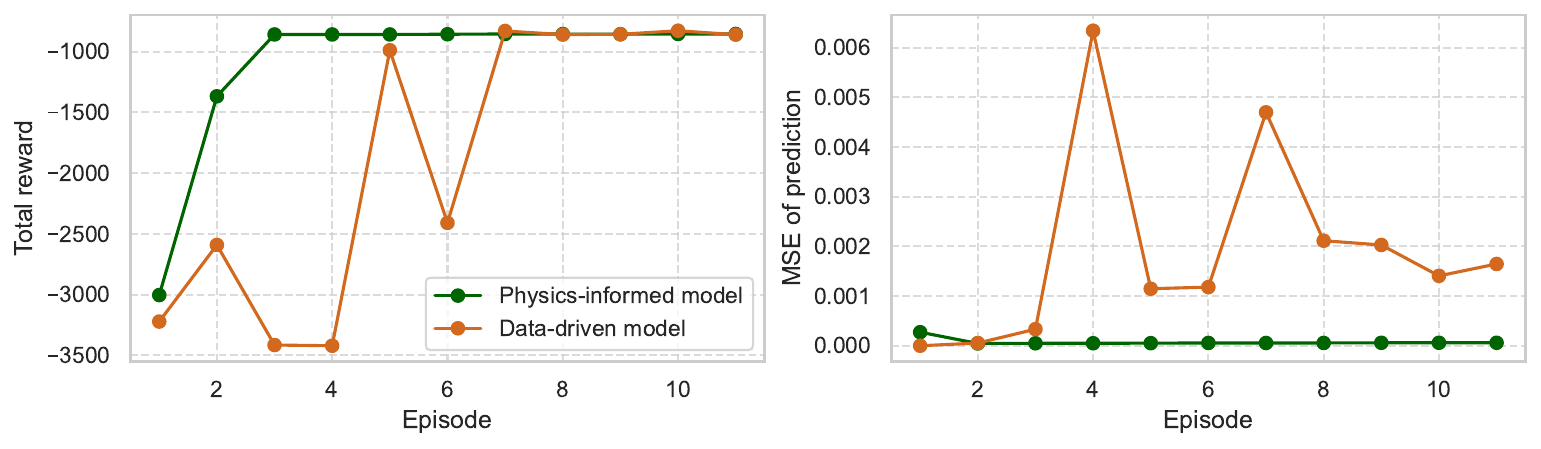}
\caption{A data-driven model still poorly predicts the next states even when its asymptotic performance matches that of the physics-informed model. Figure obtained with 10 episodes of model training on Pendulum swingup.}
\label{figure:generalization}
\end{figure}
Actually, learning a policy and Q-function through imagination leads to superior performance only when the model used to generate samples is accurate enough. 
Figure 5 in Appendix D.3 shows that an agent trained on imaginary trajectories generated with a physics-informed model largely outperforms the same agent using a fully data-driven model and matches the performance of TD3 which is trained on real trajectories.
This highlights the capability of the physics-informed model to immediately generalize to unseen data, in contrast to the data-driven model, which poorly predicts trajectories in unseen states. \cref{figure:generalization} illustrates this faster generalization capability, showing that the agent with a data-driven model still poorly predicts trajectories even when it meets the asymptotic performance of the agent with the physics-informed model.

\section{Conclusion}

We have introduced PhiHP, a novel approach that leverages physics knowledge of system dynamics to address the trade-off between asymptotic performance, sample efficiency, and time efficiency in RL. PhIHP enhances the sample efficiency by learning a physics-informed model that serves to train a model-free agent through imagination and uses a hybrid planning strategy to improve the inference time and the asymptotic performance. In the future, we envision to apply PhIHP to more challenging control tasks where there is a larger discrepancy between the known equations and the real dynamics of the system.


\bibliography{main}
\bibliographystyle{rlc}


\appendix
\newpage
\section{Comparison to existing methods}

\begin{figure}[h]
\begin{center}
\centerline{\includegraphics[width=0.3\linewidth]{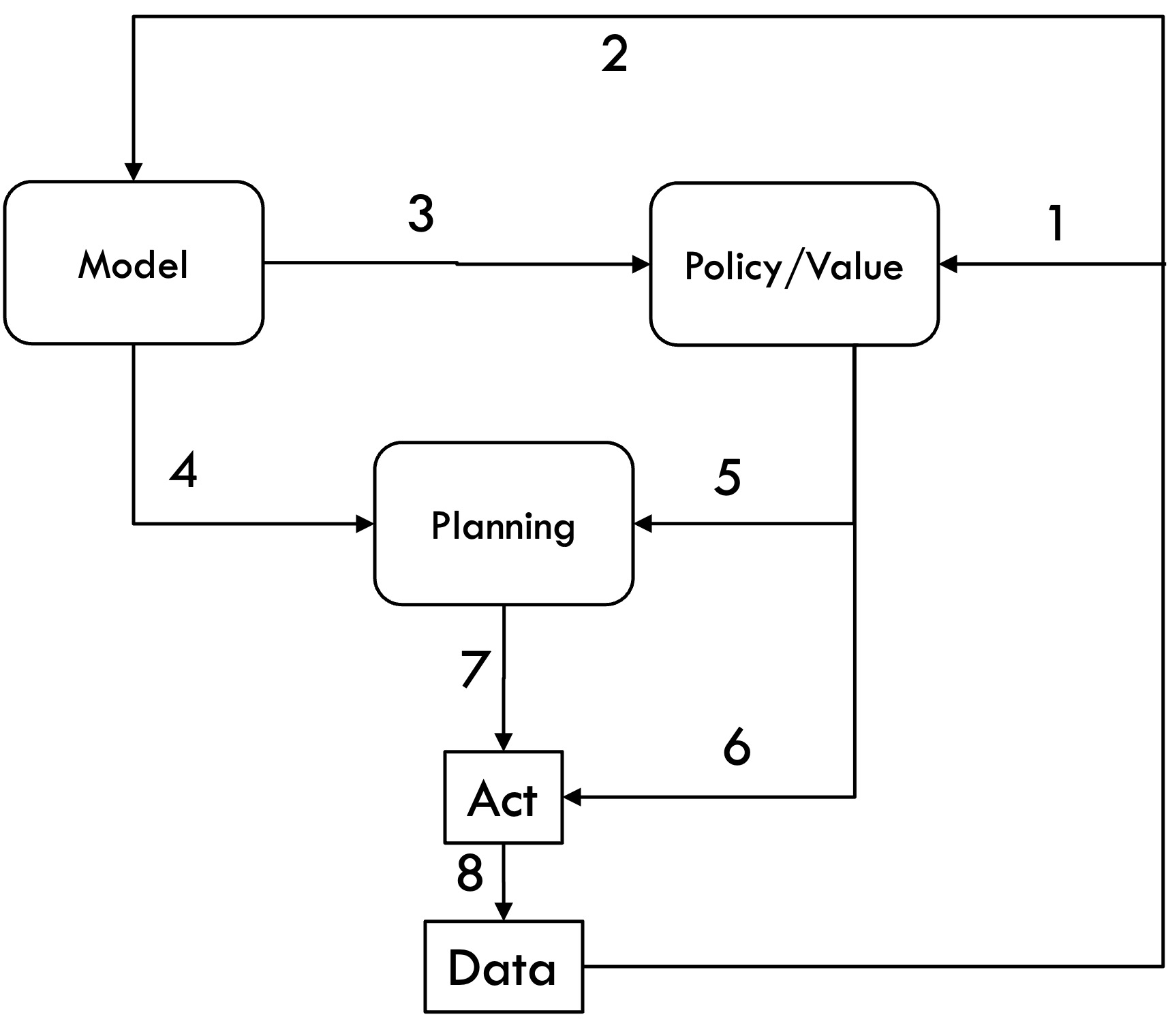} }
\caption{Overview of existing scheme of learning/planning in RL. 1- learn a policy/value function from real data. 2- learn a model from real data. 3- learn a policy/value function from imaginary data. 4- plan with a learned model. 5- plan with a learned policy/value function.  6- act based on a policy output. 7- act based on the planning outcome. 8- collect data from the interaction with the real world.}
\label{fig:generalRL}
\end{center}
\end{figure}

\begin{figure}[h]
    \begin{subfigure}{0.28\textwidth}
        \centering
        \includegraphics[width=\linewidth]{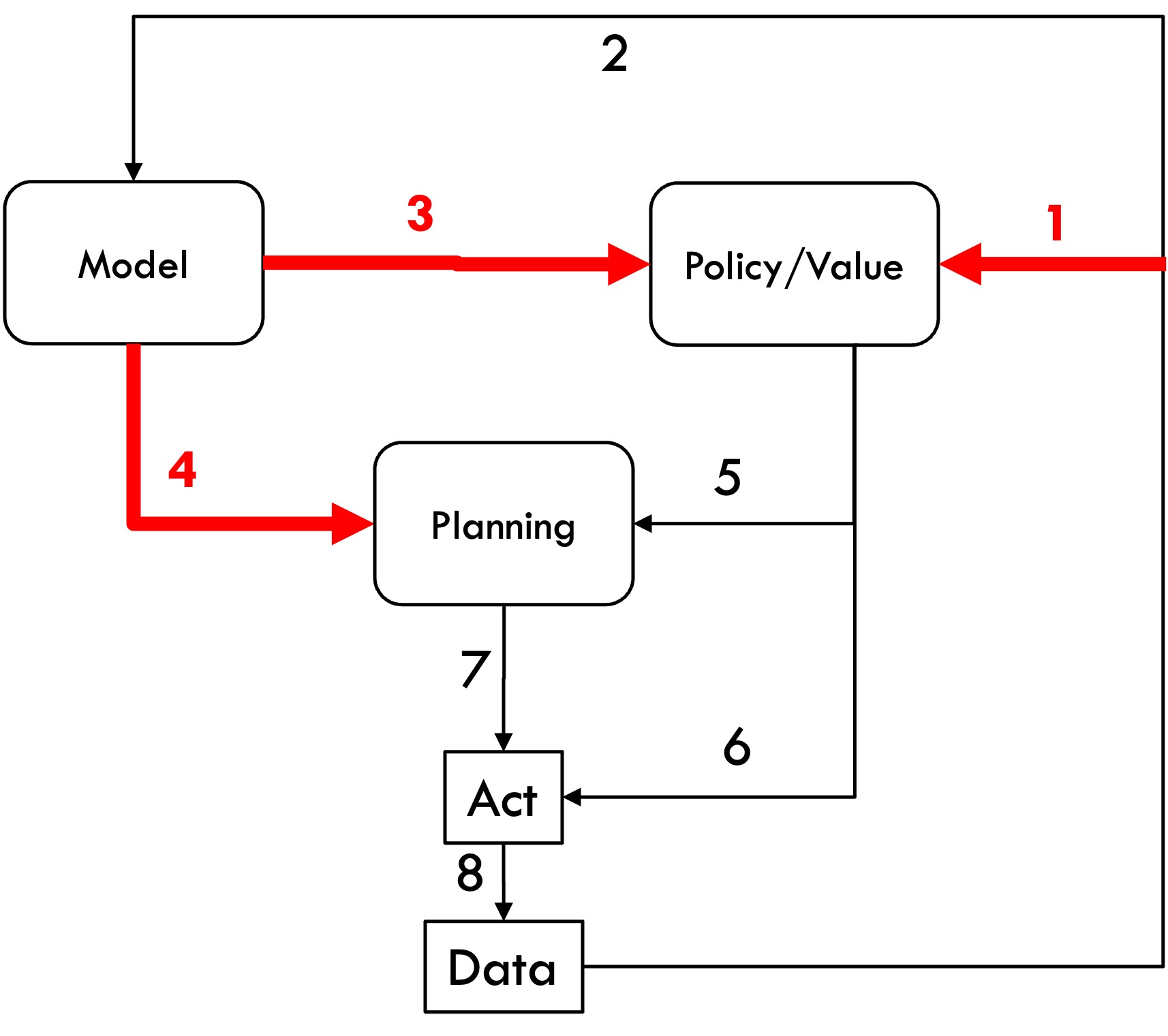}
        \caption{general scheme}
    \end{subfigure}
    \hfill
    \begin{subfigure}{0.28\textwidth}
        \centering
        \includegraphics[width=\linewidth]{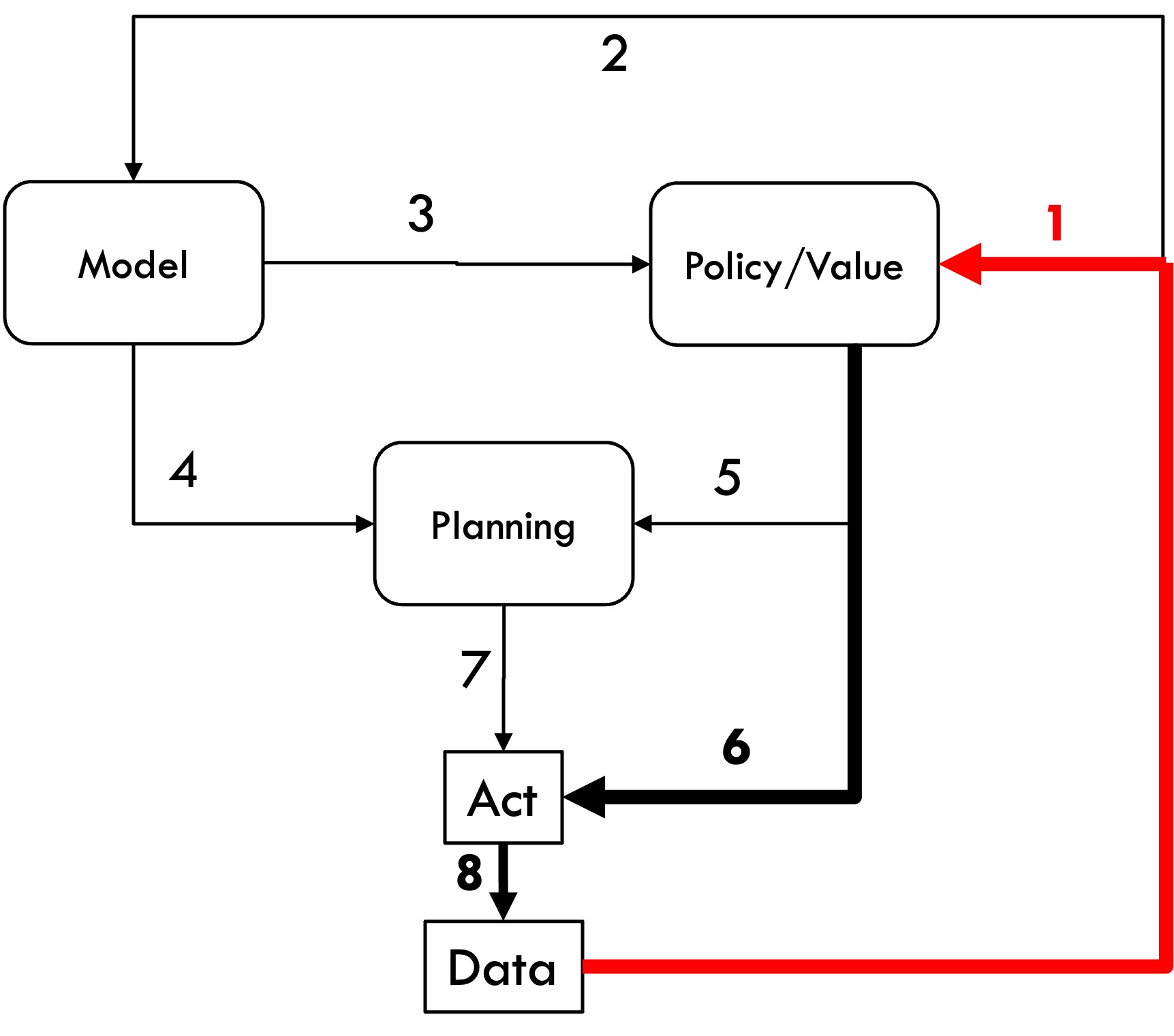}
        \caption{MFRL (TD3, SAC)}
    \end{subfigure}
    \hfill    
    \begin{subfigure}{0.28\textwidth}
        \centering
        \includegraphics[width=\linewidth]{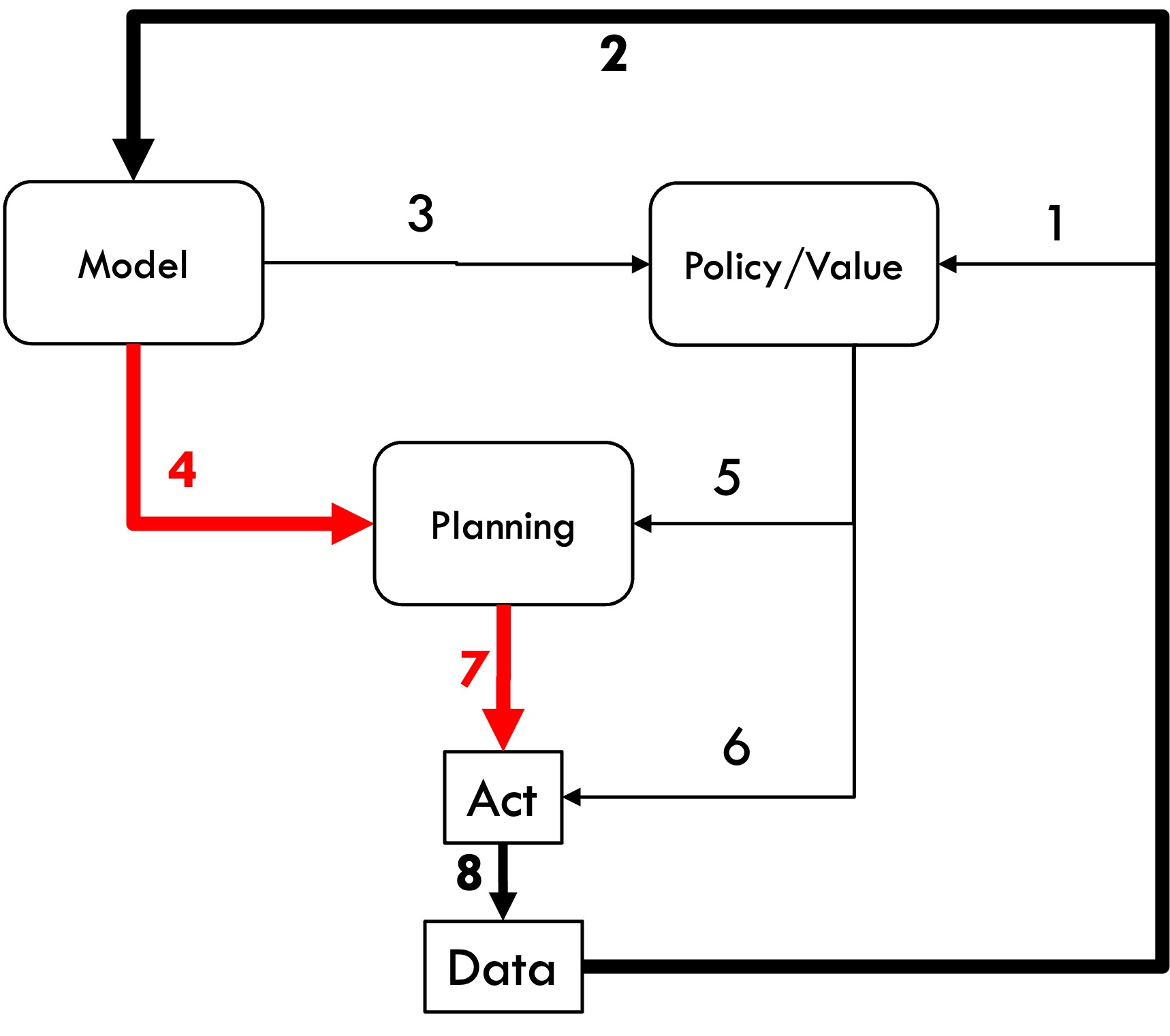}
        \caption{MBRL (PILCO)}

    \end{subfigure}

\medskip

    \begin{subfigure}{0.28\textwidth}
        \centering
        \includegraphics[width=\linewidth]{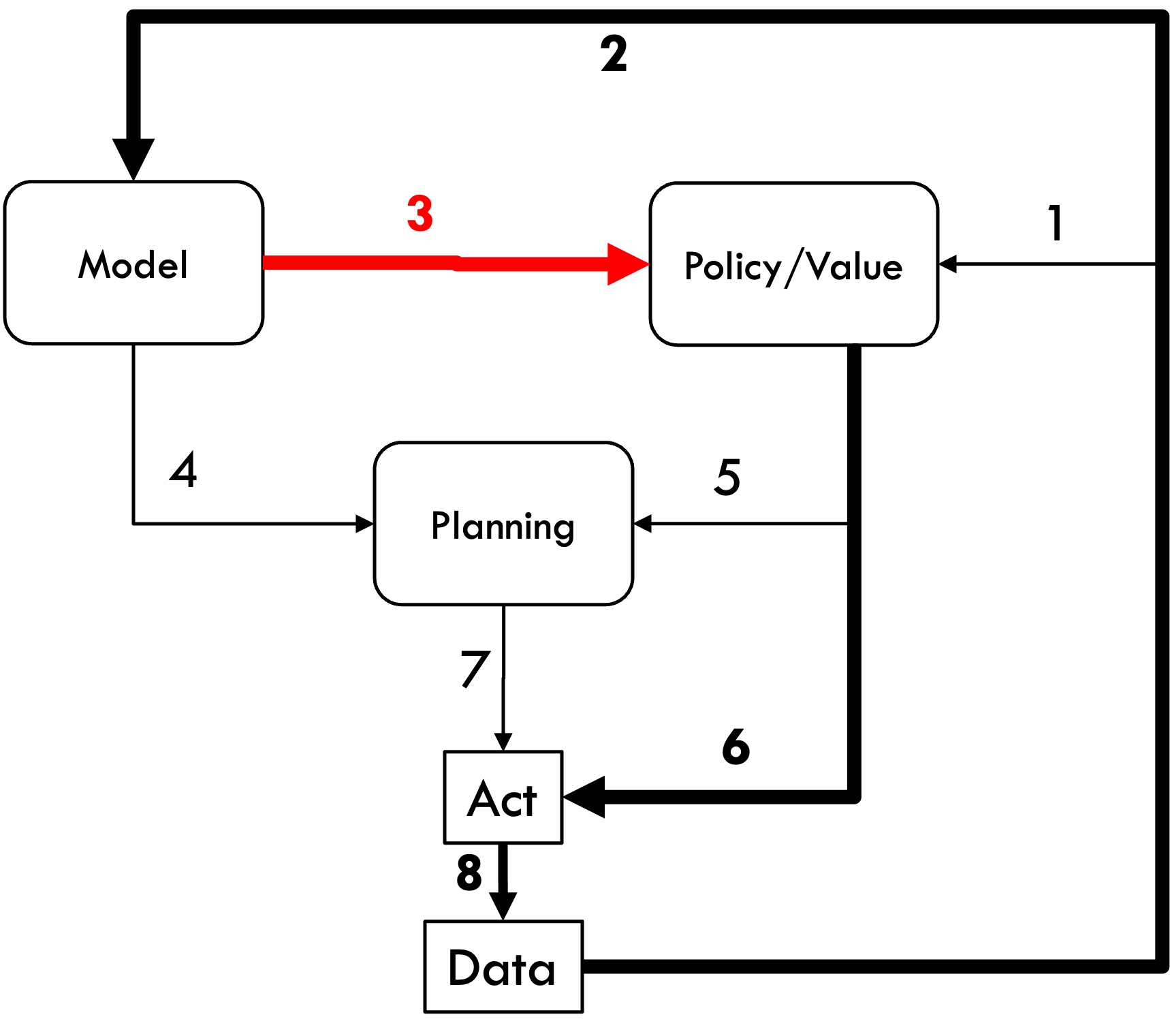}
        \caption{Dyna-style RL (LOOP)}

    \end{subfigure}
    \hfill    
    \begin{subfigure}{0.28\textwidth}
        \centering
        \includegraphics[width=\linewidth]{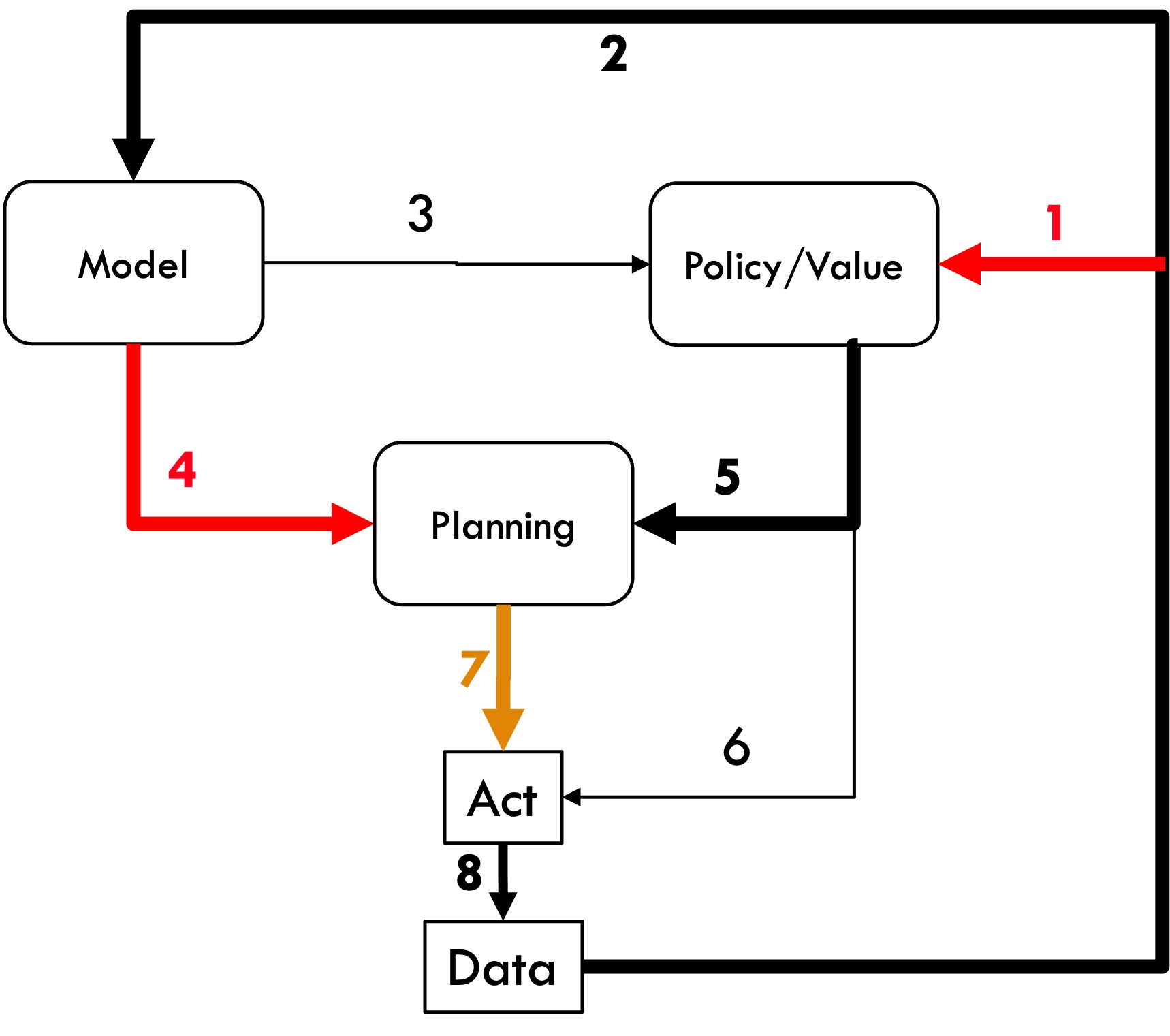}
        \caption{Hybrid RL (TD-MPC)}

    \end{subfigure}
    \hfill
    \begin{subfigure}{0.28\textwidth}
        \centering
        \includegraphics[width=\linewidth]{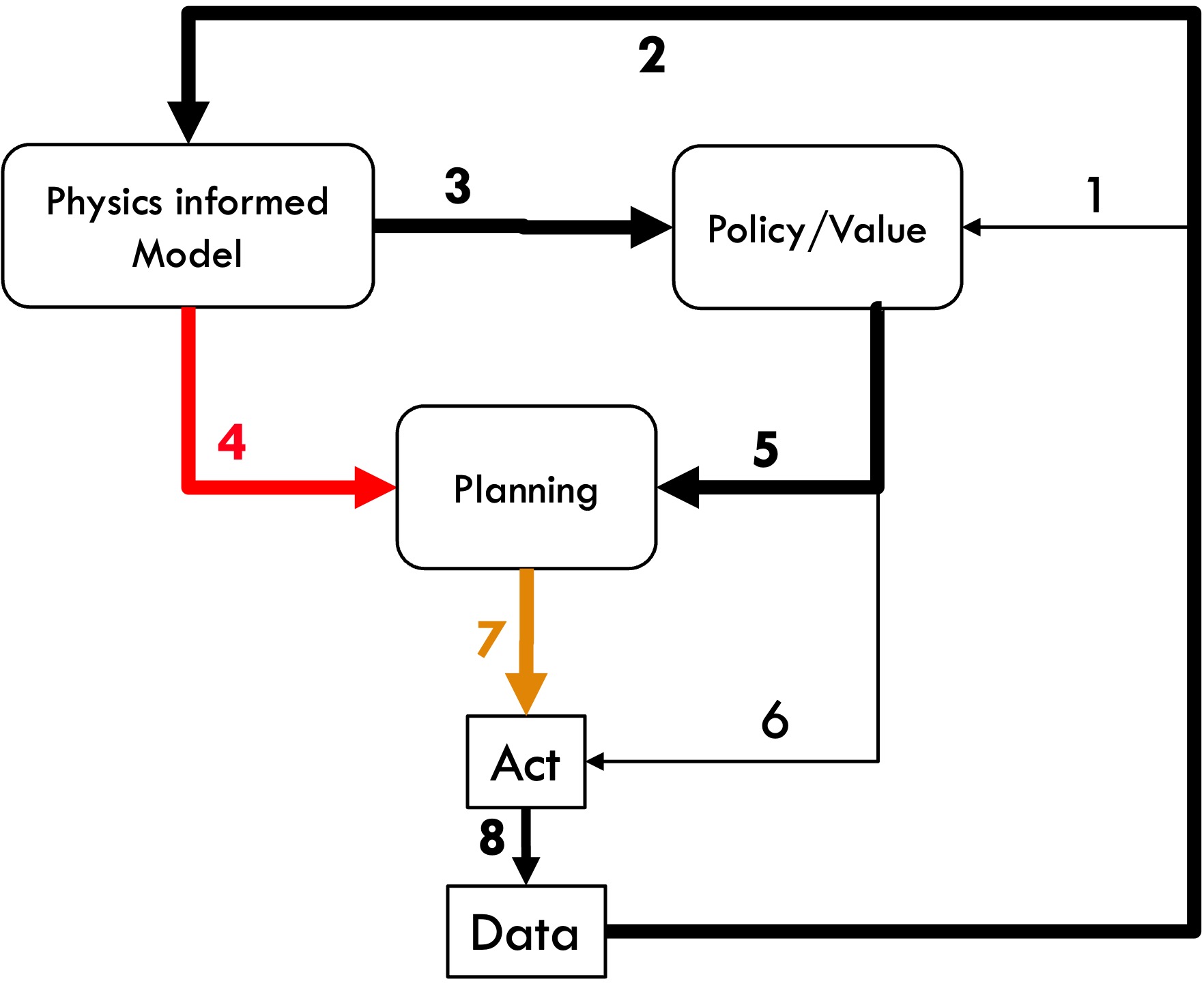}
        \caption{PhIHP (Ours)}

    \end{subfigure}

\medskip

\caption{Conceptual comparison of PhIHP and existing methods based on the general scheme in \cref{fig:generalRL}. Thick lines are used by a method, red lines indicate the origin of the main drawbacks: 1- learning on real data impacts \textbf{the sample efficiency}, 3- bias introduced by the data-driven model impacts \textbf{the asymptotic performance}, 4- planning with a model impacts the \textbf{inference time}. 
}
    \label{fig:comparison}
\end{figure}

In this section, we present a conceptual comparison of PhIHP and existing RL methods. \cref{fig:generalRL} illustrates the general scheme of existing RL methods and the possible connections between learning and planning. We highlight in \cref{fig:comparison} the origin of the well-known drawbacks in RL: $i)$ learning a policy on real data (arrow 1) impacts the sample efficiency, $ii)$ learning a policy from a data-driven learned model (arrow 3) impacts the asymptotic performance due to the bias in the learned model, $iii)$ model-based planning (arrow 4) impacts the inference time.

PhIHP benefits from the good sample efficiency of model-based learning methods (arrow 2) and from the physical knowledge to reduce the bias in the learned model. The accurately learned model generates good trajectories to train the policy/value networks (arrow 3). When interacting with the environment, PhIHP uses a hybrid planning strategy (arrows 4 \& 5) to improve asymptotic performance and time efficiency.
\section{Environments}
\label{app_env}

\begin{figure}[ht]
\begin{center}
\centerline{\includegraphics[width=\linewidth]{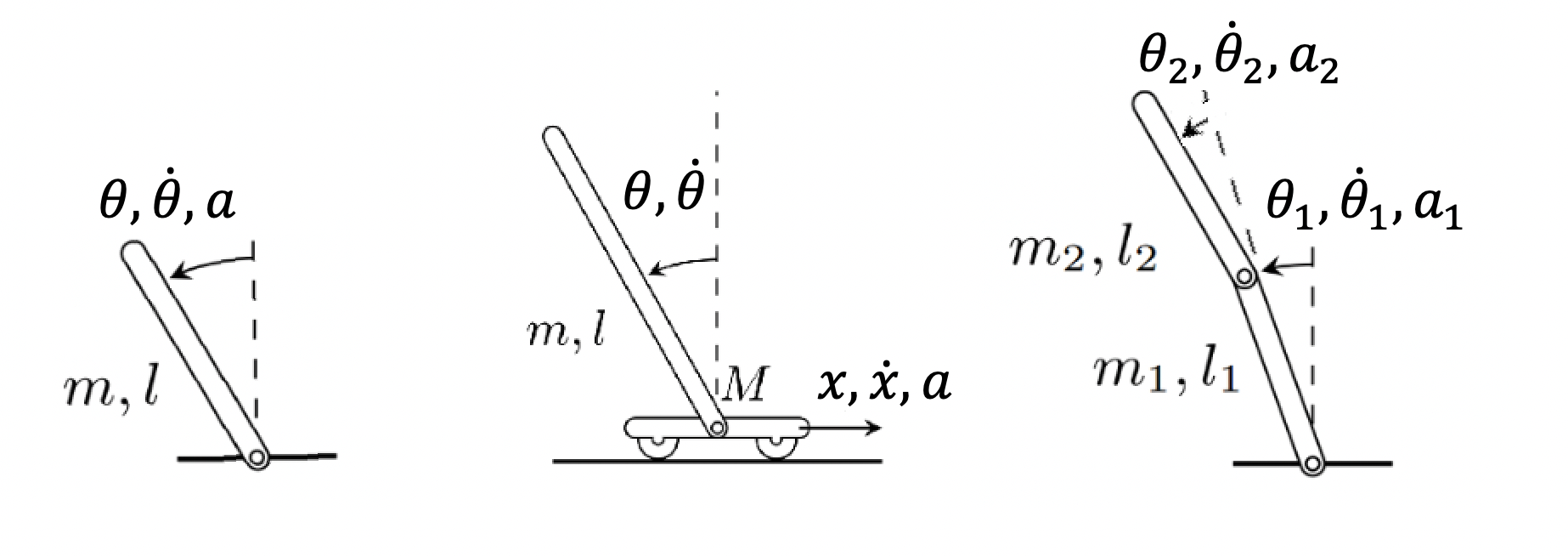} }
\caption{Experimental tasks : Pendulum \& Pendulum-swingup (left), Cartpole \& Cartpole-swingup (center), Acrobot \& Acrobot-swingup(right). The Acrobot-swingup is fully actuated while Acrobot is only actuated at the joint between the two links, thus $a_2 = 0$.}
\label{fig:tasks}
\end{center}
\end{figure}

In this section, we give a comprehensive description of the environments employed in our work. Across all environments, observations are continuous within $\left[ -S_{box} , S_{box} \right]$ and actions are continuous and restricted
to a $\left[ -a_{max} , a_{max} \right]$ range. An overview of all tasks is depicted in \cref{fig:tasks} and 
specific parameters are outlined in Table~\ref{tab:specifications-table-commun}.

\textbf{Pendulum:} A single-linked pendulum is fixed on one end, with an actuator on the joint. The pendulum starts at a random position and the goal is to swing it up and balance it at the upright position. Let $\theta$ be the joint angle at time $t$ and $\dot \theta$ its velocity, the observation at time $t$ is $(\theta, \dot \theta)$. 

\textbf{Pendulum-Swingup:} the version of Pendulum where it is started at the "hanging down" position. 

\textbf{Cartpole:} A pole is attached by an unactuated joint to a cart, which moves along a horizontal track. The pole is started upright on the cart and the goal is to balance the pole by applying forces in the left and right direction on the cart.

\textbf{Cartpole-Swingup:} the version of Cartpole where the pole is started at the "hanging down" position.

\textbf{Acrobot:} A pendulum with two links connected linearly to form a chain, with one end of the chain fixed. Only the joint between the two links is actuated. The goal is to apply torques on the actuated joint to swing the free end of the linear chain above a given height.

\textbf{Acrobot-Swingup:} For the swingup task, we experiment with the fully actuated version of the Acrobot similarly to \citep{yildiz2021continuous, xie2016model}. Initially, both links point downwards at the "hanging down" position. The goal is to swing up the Acrobot and balance it in the upright position. Let ${\theta_1}$ be the joint angles of the first fixed to a hinge at time $t$ and ${\theta_2}$ the relative angle between the two links at time $t$.
The observation at time $t$ is $({\theta_1}, {\theta_2}, \dot {\theta_1}, \dot {\theta_2})$.

\begin{table}[ht]
\centering
\resizebox{\textwidth}{!}{
\begin{tabular}{|l|c|c|c|c|c|c|c|r|}
    \toprule

    & \multicolumn{6}{c|}{Environments}                   \\
    \cmidrule(r){2-7}
    Parameters     & Pendulum & Pendulum-SU    & Cartpole & Cartpole-SU & Acrobot & Acrobot-SU \\
    \midrule
    Reward type & Smooth & Smooth    & Sparse & Smooth & Sparse & Smooth \\    
    Early termination & No & No    & Yes & No & Yes & No \\       
    \midrule
    State space &  \multicolumn{2}{c|}{$ \mathbb{R}^2$}      & \multicolumn{2}{c|}{$ \mathbb{R}^4$}  & \multicolumn{2}{c|}{$ \mathbb{R}^4$}    \\  
    States & \multicolumn{2}{c|}{$\left[ { \theta } , \dot {\theta} \right]$} & \multicolumn{2}{c|}{$\left[ x, \dot x, { \theta } , \dot {\theta} \right]$} & \multicolumn{2}{c|}{$\left[ \theta_1, \theta_2, \dot {\theta_1} , \dot {\theta_2}  \right]$ }   \\      
    \midrule
    
    Observation space & \multicolumn{2}{c|}{$ \mathbb{R}^3$}      & \multicolumn{2}{c|}{$ \mathbb{R}^5$}  & \multicolumn{2}{c|}{$ \mathbb{R}^6$}    \\    
    Observations & \multicolumn{2}{c|}{$\left[ { cos(\theta) } , { sin(\theta) }, \dot {\theta} \right]$} & \multicolumn{2}{c|}{$\left[ x, \dot x, { cos(\theta) } , { sin(\theta) } , \dot {\theta} \right]$} & \multicolumn{2}{c|}{$\left[ cos(\theta_1), sin(\theta_1),cos(\theta_2),sin(\theta_2), \dot {\theta_1} , \dot {\theta_2}  \right]$ }   \\   

    \midrule

    Actions space    & $ \mathbb{R}^1$      & $ \mathbb{R}^1$ & $ \mathbb{R}^1$ & $ \mathbb{R}^1$ & $ \mathbb{R}^1$ & $ \mathbb{R}^2$   \\
    $a_{max}$     & $\left[ 2.0 \right]$      & $\left[ 2.0 \right]$ & $\left[ 10.0 \right]$ & $\left[ 10.0 \right]$ & $\left[ 1.0 \right]$ & $\left[ 1.0 , 1.0 \right]$   \\
    \midrule      
    Length of the rollout    & 200    & 500  & 500& 500& 500& 500   \\
    \midrule    
    $\Delta t$ & \multicolumn{2}{c|}{ 0.05} & \multicolumn{2}{c|}{ 0.02} & \multicolumn{2}{c|}{ 0.2}     \\

    \bottomrule

\end{tabular}
}    
\caption{Environment specifications}
\label{tab:specifications-table-commun}
\end{table}

\subsection{Dynamic functions}

In this section, we provide details of the dynamic functions. For each task, the dynamic function consists of a frictionless component and a friction term.

\textbf{Pendulum and Pendulum Swingup:} Let $s_t = (\theta, \dot \theta)$ be the state and $a_t$ the action at time $t$. The dynamic of the pendulum is described as: 
\begin{equation}
    \text{ } 
    F(\V s_t,  a_t)=
     \begin{bmatrix}
        \dot \theta  \\
        \ddot \theta  
     \end{bmatrix} =      
     \begin{bmatrix}
        \dot \theta  \\
        C_{g} \cdot sin(\theta) + C_{i} \cdot a_t + C_{Fr} \cdot \dot \theta 
     \end{bmatrix}
     \label{eq:pendulum_2}
\end{equation}

where $C_{g}$ is the gravity norm,  $C_{i}$ is the inertia norm and $C_{Fr}$ is the Friction norm.

\textbf{Acrobot and Acrobot Swingup:} Let $s_t = (\theta_1,\theta_2, {\dot \theta}_1, {\dot \theta}_2)$ be the state and $a_t = (a_1, a_2)$ ($a_1 = 0$ for the Acrobot environment) the action at time $t$. The dynamic of the system is similar to \citep{yildiz2021continuous}  described as:

\begin{equation}
    \text{ } 
    F(\V s_t,  a_t)=
     \begin{bmatrix}
        \dot \theta_1  \\
        \dot \theta_2  \\    
        \ddot \theta_1  \\
        \ddot \theta_2  
     \end{bmatrix}    = 
         \begin{bmatrix}
        \dot \theta_1  \\
        \dot \theta_2  \\
        \frac{-(\alpha_0 + d_2 + \ddot \theta_2 + \Sigma1)}{ d_1} \\
        \frac{\alpha_1 + \frac{d_2}{d_1} \cdot \Sigma_1 - m_2 \times l_1 \cdot lc_2 \times {\dot \theta_1}^2 \cdot \sin{\theta_2} - \Sigma_2}{m2 \cdot {lc_2}^2 + I_2 - \frac{{d_2}^2}{d_1}} 
     \end{bmatrix}
    \text{ }
     \label{eq:acrobot}
\end{equation}

where: \\
$  ~~~ ~~~ ~~~ \alpha_0 = a_1 - C_{fr1} \cdot \dot \theta_1 \color{black}\text{ such as }  C_{fr1} \text{ is the friction norm in the first joint }$, \\
$ ~~~ ~~~ ~~~ \alpha_1 = a_2 - C_{fr2} \cdot \dot \theta_2 \color{black}\text{ such as }  C_{fr2} \text{ is the friction norm in the second joint }$, \\
$ ~~~ ~~~ ~~~ m_1$ and $m_2$ the mass of the first and second links, \\
$ ~~~ ~~~ ~~~ l_1$ and $l_2$ the length of the first and second links, \\
$ ~~~ ~~~ ~~~ lc_1$ and $lc_2$ the position of the center of mass of the first and second links, \\
$ ~~~ ~~~ ~~~ I_1$ and $I_2$ the moment of inertia of the first and second links, \\

and

$  ~~~ ~~~ ~~~       d_1 =
            m_1 \cdot {lc_1}^2
            + m_2 \cdot ({l_1}^2 + {lc_2}^2 + 2 \cdot l_1 \cdot lc_2 \cdot \cos(\theta_2))
            + I_1
            + I_2 $

$   ~~~ ~~~ ~~~      d_2 =
            m_2 \cdot ({lc_2}^2
            +  l_1 \cdot lc_2 \cdot \cos(\theta_2))
            + I_2 $

$   ~~~ ~~~ ~~~     \Sigma_2 =
            m_2 \cdot {lc_2} \cdot g \cdot \cos(\theta_1 + \theta_2 - \frac{\pi}{2}) $

$   ~~~  ~~~~~~     \Sigma_1 =
            m_2 \cdot l_1 \cdot {lc_2} \cdot {\ddot \theta_2} \cdot \sin(\theta_2) \cdot({\ddot \theta_2} - 2   \cdot {\ddot \theta_1}) 
            + (m_1 \cdot lc_1 + m_2 \cdot l_1) \cdot g \cdot 
            \cos(\theta_1  - \frac{\pi}{2}) + \Sigma_2.$

\textbf{Cartpole and Cartpole Swingup:} Let $s_t = (\V x, \dot x, \theta, {\dot \theta})$ be the state and $a_t$ the action at time $t$. The dynamic of the system is based on \citep{barto1983neuronlike} and described as:

\begin{equation}
F(\V s_t, a_t)=
\begin{bmatrix}
\dot x \\
\ddot x \\
\dot \theta \\
\ddot \theta
\end{bmatrix} =
\begin{bmatrix}
\dot x \\
\Sigma - m_p \cdot l \cdot \ddot \theta \cdot \frac{\cos(\theta)}{m_{total}} \\
\dot \theta \\
\frac{g \cdot \sin(\theta)-(\cos(\theta) \cdot \Sigma) - \frac{Fr_p \dot \theta}{m_p \cdot l}}{l \cdot [\frac{4}{3}-\frac{m_p \cdot {\cos(\theta)}^2}{m_{total}}]}
\end{bmatrix},
\label{eq:cartpole}
\end{equation}
where: \\
$~~~ ~~~ ~~~ Fr_{c} \text{ is the friction norm in the contact between the cart and the ground}$,\\
$~~~ ~~~ ~~~ Fr_{p} \text{ is the friction norm in the joint between the cart and the pole}$, \\
$~~~ ~~~ ~~~ l$ is the length of the pole, \\
$~~~ ~~~ ~~~ m_{tot}=m_c+m_p$ and $m_p$, $m_c$ are the mass of the pole and the cart respectively, \\
$~~~ ~~~ ~~~ \Sigma = \frac{1}{m_{total}} \cdot (a + m_p \cdot l \cdot {\dot \theta}^2 \cdot \sin(\theta) -(  Fr_c \cdot \text{sgn}(\dot x) )$.

\subsection{Reward Functions}
The reward function encodes the desired task. We adopt the original reward functions in the three main environments. For the swingup variants, we choose functions that describe the swingup task: we adopt the same function as Pendulum for Pendulum swingup. For Cartpole swingup, we set a reward function as the negative distance from the goal position ${s}_{goal} = (x=0, y=1)$. For Acrobot swingup, we take the height of the pole as a reward function.

\begin{table}[ht]
\begin{center}
\begin{small}
\centering
\resizebox{0.5\textwidth}{!}{
\begin{tabular}{|l|c|}
    \hline
    Environment & Reward function      \\
    \hline
    Pendulum     &  $-{\theta}^2 - 0.1 \cdot \dot{\theta}^2 -0.001\cdot {a}^2$\\
    Pendulum swingup    &  $-{\theta}^2 - 0.1 \cdot \dot{\theta}^2 -0.001\cdot {a}^2$ \\
    Cartpole    & $+1$ for every step until termination\\
    Cartpole swingup & $\exp{(\Vert \V s - {s}_{goal} \Vert ^2_2 )}$ \\
    Acrobot & -1 for every step until termination  \\
    Acrobot swingup  & $-\cos({\theta}_1) - \cos({\theta}_1 + {\theta}_2)$\\    
    \hline

\end{tabular}
}    

\end{small}
\end{center}
\caption{Reward functions for each environment.}
\label{table:reward}
\end{table}

\section{Implementation details}
\label{app_imp}
In this section, we describe the experimental setup and the implementation details of PhIHP. We first learn a physics-informed residual dynamics model, then learn an MFRL agent through imagination, and use a hybrid planning strategy at inference. \\
To learn the model, we first use a pure exploratory policy during $T$ timesteps to collect the initial samples to fill $\mathcal D_{re}$, then we perform stochastic gradient descent on the loss function (Eq. 3 in Sec. 4.1) to train $F_{\theta}$. The learned model $\hat{F}$ is used with CEM to perform planning and gather new $T$ samples to add to $\mathcal D_{re}$. To improve the quality of the model, the algorithm iteratively alternates between training and planning for a fixed number of iterations.

To train the model-free component of PhIHP, the training dataset $\mathcal D_{im}$ is initially filled with $T'$ samples generated from the learned model $\hat{F}$ and random actions from a pure exploratory policy, $\pi_{\theta}$ and $Q_{\theta}$ are trained on batches from $\mathcal D_{im}$ which is continuously filled by samples from the learned model $\hat{F}$.

We list in \cref{table:implementation} the relevant hyperparameters of PhIHP and baselines. and we report in \cref{table:spec_impl} the task-specific hyperparameters for PhIHP.\\
We adopted the original implementation and hyperparameters of TD-MPC. However, we needed to adapt it for early termination environments (\ie Cartpole and Acrobot) to support episodes of variable length, and we found it beneficial for TD-MPC to set the critic learning rate at 1e-4 in these two tasks. \\
Fot TD3, we tuned the original hyperparameters and used the same for the TD3 baseline and the model-free component of PhIHP.
\begin{table}[h]

\begin{center}
\begin{small}
\begin{tabular}{lccccr}
\toprule
Hyperparameter & PhIHP & TD-MPC & TD3  & CEM-oracle \\
\midrule
\multicolumn{5}{c}{Model learning}\\
\midrule
Model& ODE + MLP& MLP & - & Ground truth \\
Activation & Relu & ELU & -& - \\
MLP size & 2 x 16 & 2 x 512 & - & -\\
Learning rate& 1e-3&1e-3 & - & -\\
\midrule
\multicolumn{5}{c}{Policy/Value learning}\\
\midrule
Batch size & 64 & 512 & 64& - \\
Critic size & 3 x 200 & 2 x 512 & 3 x 200 & -\\
Actor size & 2 x 300 & 2 x 512 & 2 x 300 & - \\
 Activation & Relu & ELU & Relu & -\\
Critic learning rate  & 1e-4 & 1e-3 & 1e-4 & -\\
Actor learning rate & 1e-3 & 1e-3 & 1e-3& -\\
Soft update coefficient $\tau$ & 0.05 & 0.01& 0.05& -\\
Policy update frequency & 2 & 2& 2& -\\

Discount factor & 0.99 & 0.99 & 0.99 & - \\
Exploratory steps & 10000 & 5000 & 10000 & - \\
Replay Buffer size  & 1e6 & 1e6 & 1e6 & - \\
Sampling technique & Uniform & PER $(\alpha = 0.6, \beta = 0.4$) & Uniform & - \\

\midrule
\multicolumn{5}{c}{Planning}\\
\midrule
Planner & CEM & MPPI & -  & CEM \\
Exploratory population size  & 200 & 512 & - & 700 \\
Policy population size & 20 & 25 & - & - \\
Elite  & 10 & 64 & - & 20\\
CEM iterations $I$  & 3 & 6 & - & 3 \\
Update distribution  & mean and std. & weighted mean and std. & - & mean and std.\\
Planning horizon $H$ & 4 & 5 & -  & 30 \\
Receding horizon $RH$ & 1 & 1 & -& 5\\

\bottomrule
\end{tabular}
\caption{PhIHP and baselines hyperparameters. We emphasize that we use the same hyperparameters for TD3 in the baseline and the model-free component of PhIHP.}
\label{table:implementation}
\end{small}
\end{center}

\end{table}

\begin{table}[h]

\begin{center}
\begin{small}
\resizebox{\textwidth}{!}{
\begin{tabular}{lcccccc}
\toprule
 Hyperparameter& Pendulum & Pendulum swingup & Cartpole & Cartpole swingup & Acrobot & Acrobot swingup \\
\midrule
\multicolumn{7}{c}{Model learning}\\
\midrule
MLP size & 2 x 16 & 2 x 16  & 2 x 16  & 2 x 16 & 3 x 16 & 3 x 16 \\
Loss initial coefficient $\lambda_0$ & 1e3 & 1e3& 1e3& 1e3& 1e2& 1e3\\
Loss update coefficient ${\tau}_{ph}$ & 1e3& 1e3 & 1e5& 1e5 & 1e5& 1e5\\
Samples needed & 2000& 5000& 5000& 5000 & 5000& 15000\\

\midrule
\multicolumn{7}{c}{Planning}\\
\midrule

Planning horizon $H$ & 5 & 5 & 4 & 6 & 4 & 3 \\
Reward coefficient $\alpha$ & 1.5 & 1.5 & 0.2 & 0.03 & 0.8 & 0.8 \\
\bottomrule
\end{tabular}
}
\caption{Task-specific hyperparameters of PhIHP.}
\label{table:spec_impl}
\end{small}
\end{center}
\end{table}
\newpage
\section{Comparison to state of the art}
We compare PhIHP to baselines on individual tasks, we present both statistical results and a qualitative analysis.

\subsection{Learning curves}
\begin{figure}[h]
\vskip 0.2in
    \begin{subfigure}{0.34\linewidth}
        \centering
        \includegraphics[width=\linewidth]{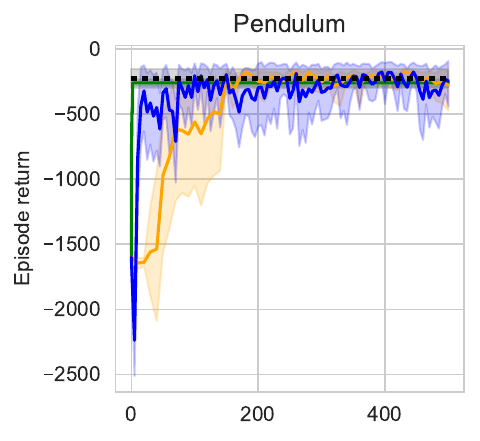}
    \end{subfigure}
    \hfill
    \begin{subfigure}{0.3\linewidth}
        \centering
        \includegraphics[width=\linewidth]{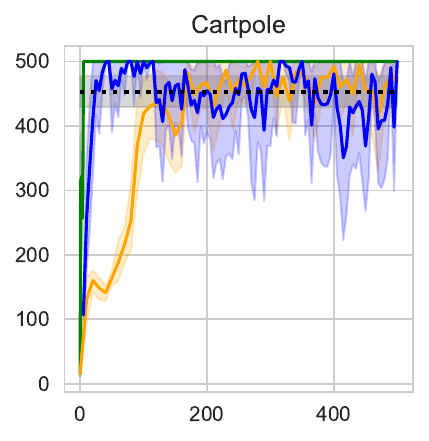}
    \end{subfigure}
    \hfill
    \begin{subfigure}{0.31\linewidth}
        \centering
        \includegraphics[width=\linewidth]{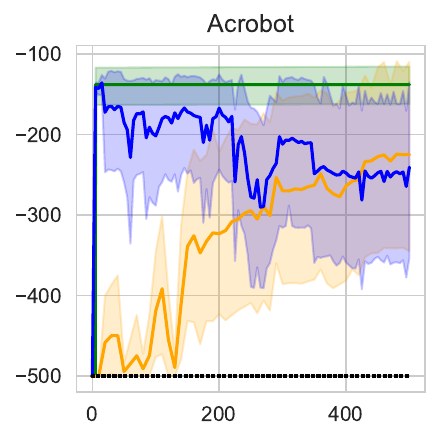}
    \end{subfigure}  
    \vfill
    \begin{subfigure}{0.34\linewidth}
        \centering
        \includegraphics[width=\linewidth]{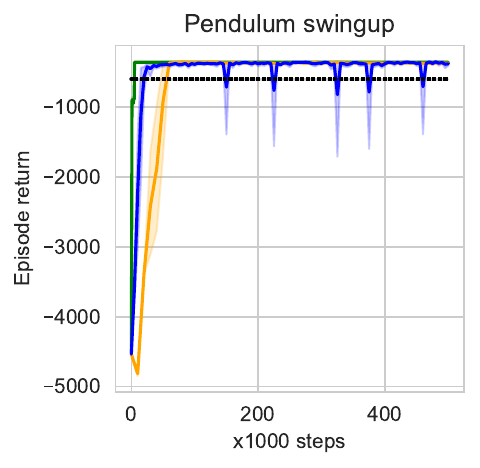}
    \end{subfigure}
    \hfill
    \begin{subfigure}{0.3\linewidth}
        \centering
        \includegraphics[width=\linewidth]{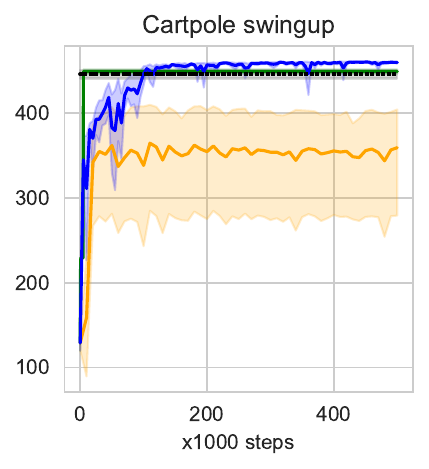}
    \end{subfigure}
    \hfill
    \begin{subfigure}{0.31\linewidth}
        \centering
        \includegraphics[width=\linewidth]{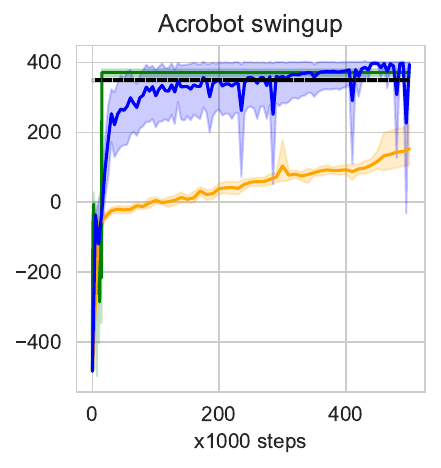}
    \end{subfigure}  

    \begin{subfigure}{\textwidth}
        \centering
        \includegraphics[width=0.8\linewidth]{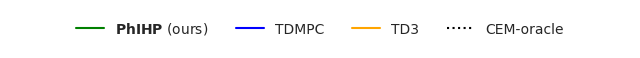}    
    \end{subfigure}

    \caption{Return of PhIHP and baselines on the gymnasium classic control tasks. Mean and std. over 10 runs. PhIHP outperforms or matches the baselines.}
    \label{fig:fig_sota_short}
\vskip -0.2in
\end{figure}

We provide learning curves of PhIHP and baselines on individual tasks. PhIHP outperforms baselines by a large margin in terms of sample efficiency.
\cref{fig:fig_sota_short} shows that TD3, even when converging early in Cartpole-swingup, achieves sub-optimal performance and fails to converge within 500k steps in Acrobot-swingup.

\subsection{Statistical Comparison: PhIHP vs. Baselines}
To ensure a robust and statistically sound comparison with the results previously reported in Table 1 in Sec. 5.2, we conducted Welch's t-test to statistically compare the performance of PhIHP \textit{vs} baselines across individual tasks. We set the significance threshold at 0.05, and calculated p-values to determine whether observed differences in performance were statistically significant. \cref{table:statistics} shows that PhIHP is equivalent to all baselines in Pendulum, and it significantly outperforms TD3 on the remaining tasks. Moreover, PhIHP outperforms TD-MPC in sparse-reward early-termination environment tasks (Cartpole and Acrobot), while they demonstrate equivalent performance in Pendulum, Pendulum swingup, and Acrobot swingup.
\begin{table}[H]
\begin{center}
\begin{small}
\resizebox{\textwidth}{!}
{
\begin{tabular}{|l||ccc||ccc||ccc|}
\toprule
 & \textbf{TD3} & \textbf{TD-MPC} & \textbf{CEM-oracle} & \textbf{TD3} & \textbf{TD-MPC} & \textbf{CEM-oracle}& \textbf{TD3} & \textbf{TD-MPC} & \textbf{CEM-oracle}  \\
\midrule
\rowcolor{gray!50}
&\multicolumn{3}{c||}{\textbf{Pendulum}} &\multicolumn{3}{c||}{\textbf{Cartpole}} &\multicolumn{3}{c|}{\textbf{Acrobot}}\\
\midrule
T-statistic & -1.41 & 0.52 & -1.18 & 4.66 & 5.47 & 25.75 & 4.39 & 5.35 & 29.78\\
P-value & 0.16 & 0.61 & 0.26 & 9.92e-06 & 3.40e-07 & 9.69e-10 & 1.96e-05 & 2.58e-07 & 3.30e-51\\
Significant difference& \textbf{No} & \textbf{No} & \textbf{No} & \textcolor{mygreen}{\textbf{Yes}} & \textcolor{mygreen}{\textbf{Yes}} & \textcolor{mygreen}{\textbf{Yes}} & \textcolor{mygreen}{\textbf{Yes}} & \textcolor{mygreen}{\textbf{Yes}} & \textcolor{mygreen}{\textbf{Yes}}\\
\rowcolor{gray!50}
&\multicolumn{3}{c||}{\textbf{Pendulum swingup}} &\multicolumn{3}{c||}{\textbf{Cartpole swingup}} &\multicolumn{3}{c|}{\textbf{Acrobot swingup}}\\
\midrule
T-statistic  & 6.35 & 1.19 & 6.47 & 8.41 & -7.59 & 1.65 & 27.49 & -0.10 & 4.02\\
P-value & 1.48e-09 & 0.24 & 1.15e-4 & 2.70e-13 & 9.01e-12 & 0.11 & 3.54e-66 & 0.92 & 1.09e-4\\
Significant difference& \textcolor{mygreen}{\textbf{Yes}} & \textbf{No} & \textcolor{mygreen}{\textbf{Yes}} & \textcolor{mygreen}{\textbf{Yes}} & \textcolor{myred}{\textbf{Yes}} & \textbf{No} & \textcolor{mygreen}{\textbf{Yes}} & \textbf{No} & \textcolor{mygreen}{\textbf{Yes}}\\
\bottomrule
\end{tabular}
}
\end{small}
\end{center}
\caption{Statistical Comparison of PhIHP \textit{vs} Baselines across individual tasks: we present the Welch's t-test results including T-statistics and P-values, to assess the significance of performance differences. \textbf{Yes} denotes a statistically significant difference ($\text{p-value} < 0.05$), with green \textcolor{mygreen}{\textbf{Yes}} indicating PhIHP outperforming the baseline ($\text{T-statistics} > 0$), and red \textcolor{myred}{\textbf{Yes}} indicating the baseline performing better ($\text{T-statistics} < 0$). \textbf{No} indicates no significant difference between PhIHP and the baseline ($\text{p-value} > 0.05$).}
\label{table:statistics}
\end{table}

\subsection{Imagination learning for model-free TD3}
\begin{figure}[H]
    \begin{subfigure}{0.33\linewidth}
        \centering
        \includegraphics[width=\linewidth]{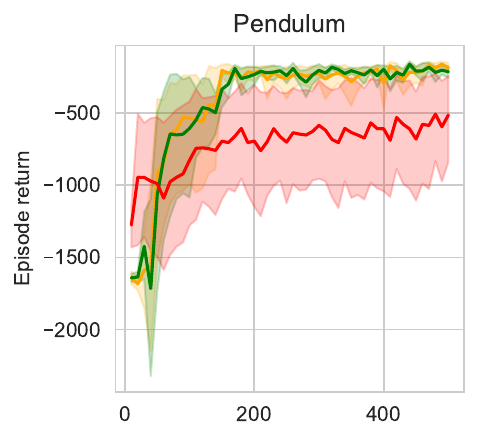}
    \end{subfigure}
    \hfill
    \begin{subfigure}{0.3\linewidth}
        \centering
        \includegraphics[width=\linewidth]{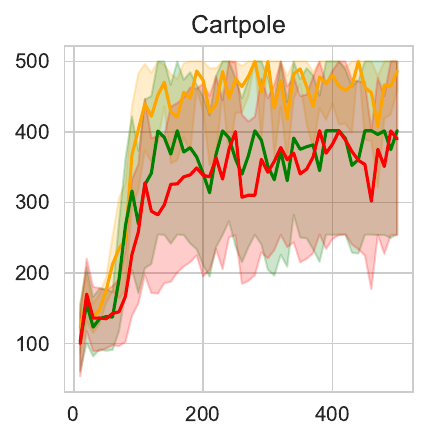}
    \end{subfigure}
    \hfill
    \begin{subfigure}{0.31\linewidth}
        \centering
        \includegraphics[width=\linewidth]{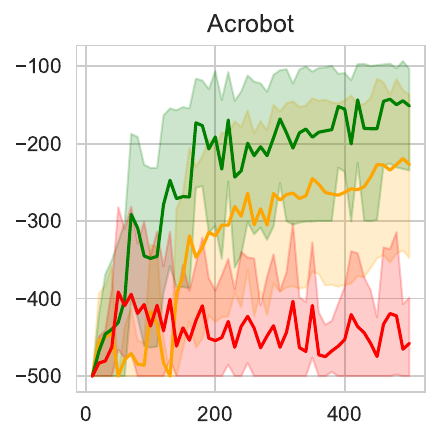}
    \end{subfigure}  
    
    \begin{subfigure}{0.33\linewidth}
        \centering
        \includegraphics[width=\linewidth]{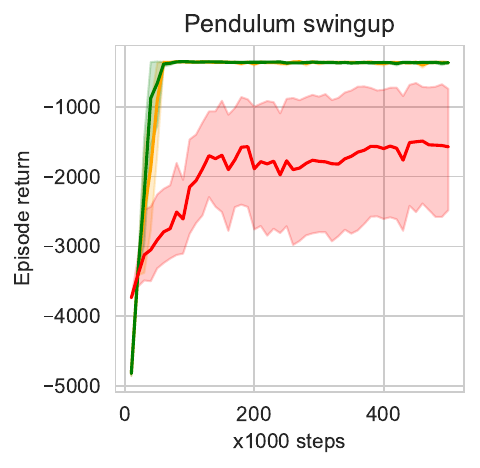}
    \end{subfigure}
    \hfill
    \begin{subfigure}{0.3\linewidth}
        \centering
        \includegraphics[width=\linewidth]{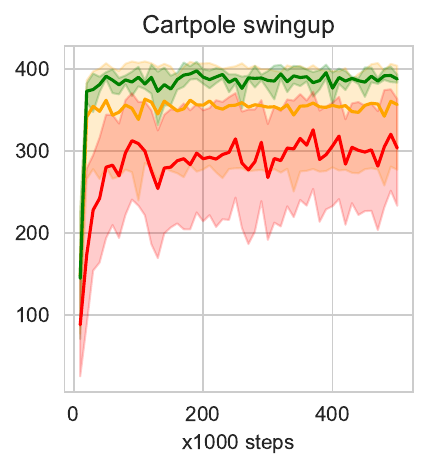}
    \end{subfigure}
    \hfill
    \begin{subfigure}{0.31\linewidth}
        \centering
        \includegraphics[width=\linewidth]{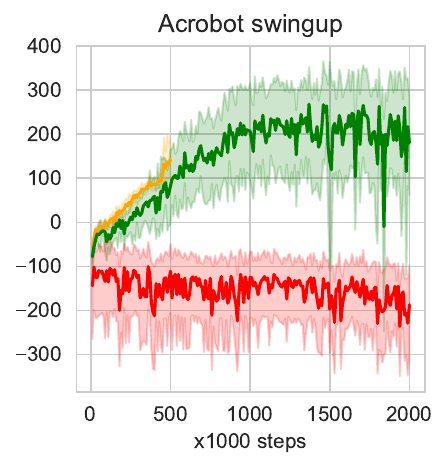}
    \end{subfigure}  
    \begin{subfigure}{\textwidth}
        \centering
        \includegraphics[width=0.6\linewidth]{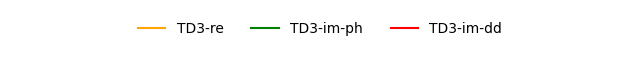}    
    \end{subfigure}
    \caption{Learning curve of TD3 on classic control tasks, mean and std. over 5 runs. \textbf{TD3-re} (orange curve) is a TD3 agent trained on real trajectories, \textbf{TD3-im-ph} (green curve) and \textbf{TD3-im-dd} (red curve) are TD3 agents trained on imaginary trajectories respectively from a physics-informed model and data-driven model. }
    \label{fig:td3_imagination}
\end{figure}

We provide learning curves of TD3 through imagination on individual tasks in \cref{fig:td3_imagination}. TD3-im-ph is a component of PhIHP, it is a TD3 agent learned on trajectories from a physics-informed model. It largely outperforms TD3-im-dd, a TD3 learned on trajectories from a data-driven model.
we limited the training budget for TD3-re, trained on real trajectories, at 500k real samples in all tasks.

\subsection{Qualitative comparison}

\begin{figure}[H]
    \begin{subfigure}{\linewidth}
        \centering
        \includegraphics[width=\textwidth]{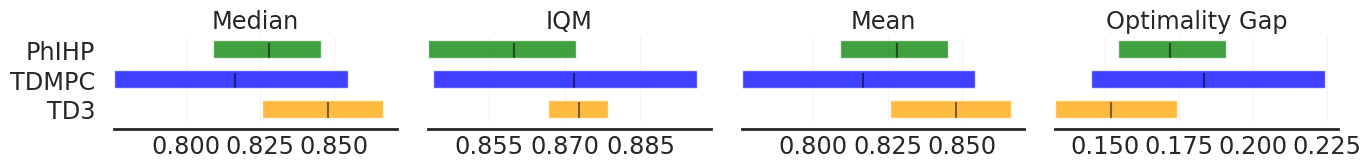}
        \caption{\textbf{Pendulum}. PhIHP matches the performance of TD-MPC and TD3.}
    \end{subfigure}
    \vfill

    \begin{subfigure}{\linewidth}
        \centering
        \includegraphics[width=\textwidth]{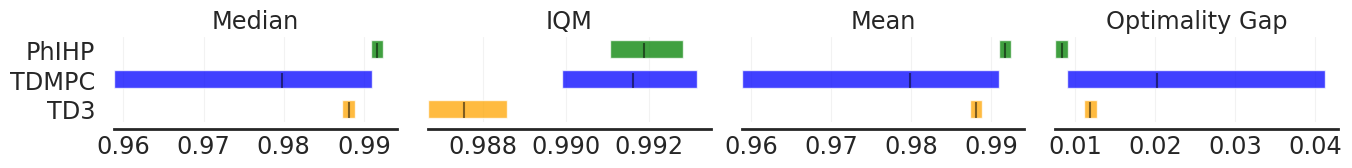}
        \caption{\textbf{Pendulum swingup}. PhIHP outperforms TD-MPC and TD3, and PhIHP shows to be robust to outliers compared to TD-MPC.}
    \end{subfigure}
    \vfill

    \begin{subfigure}{\linewidth}
        \centering
        \includegraphics[width=\textwidth]{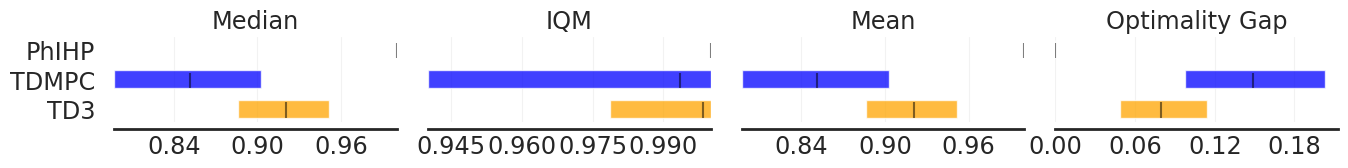}
        \caption{\textbf{Cartpole}. PhIHP largely outperforms TD-MPC and TD3.}
    \end{subfigure}
    \vfill

    \begin{subfigure}{\linewidth}
        \centering
        \includegraphics[width=\textwidth]{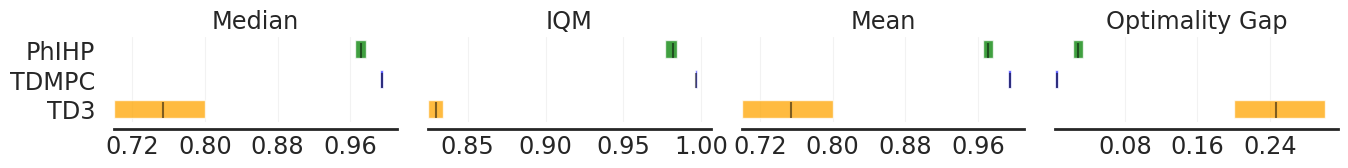}
        \caption{\textbf{Cartpole swingup}. PhIHP outperforms TD3 and shows slightly less performance than TD-MPC.}
    \end{subfigure}
    \vfill

    \begin{subfigure}{\linewidth}
        \centering
        \includegraphics[width=\textwidth]{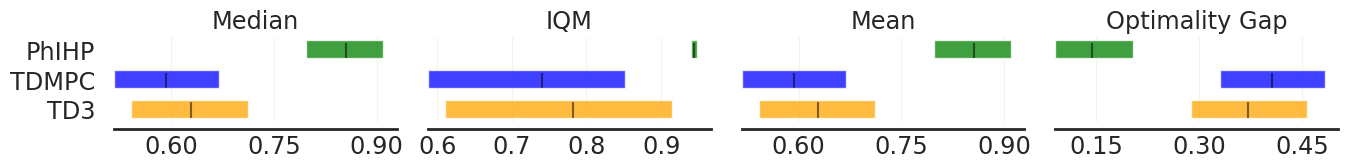}
        \caption{\textbf{Acrobot}. PhIHP largely outperforms TD3 and TD-MPC.}
    \end{subfigure}
    \vfill

    \begin{subfigure}{\linewidth}
        \centering
        \includegraphics[width=\textwidth]{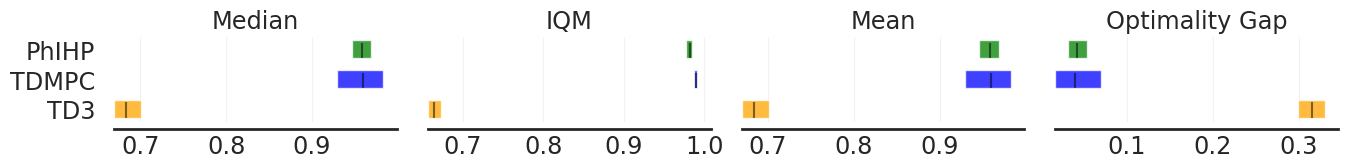}
        \caption{\textbf{Acrobot swingup}. PhIHP outperforms TD3 and matches the performance of TD-MPC.}
    \end{subfigure}
    \vfill
    \caption{Median, interquartile median (IQM), mean performance, and optimality gap of PhIHP and baselines on individual classic control tasks (10 runs). Higher mean, median, and IQM performance and lower optimality gap are better. Confidence intervals (CIs) are estimated using the percentile bootstrap with stratified sampling \citep{agarwal2021deep}.}
    \label{fig:rliable_all}
\end{figure}

\begin{figure}[H]
    \begin{subfigure}{0.33\linewidth}
        \centering
        \includegraphics[width=\linewidth]{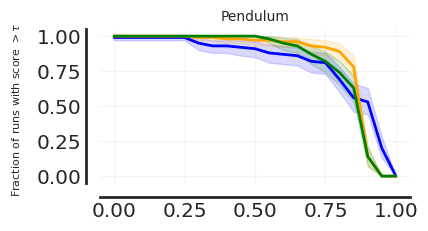}
    \end{subfigure}
    \hfill
    \begin{subfigure}{0.3\linewidth}
        \centering
        \includegraphics[width=\linewidth]{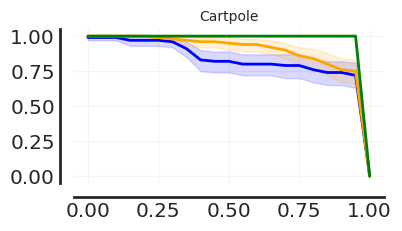}
    \end{subfigure}
    \hfill
    \begin{subfigure}{0.3\linewidth}
        \centering
        \includegraphics[width=\linewidth]{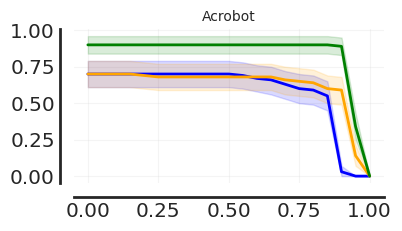}
    \end{subfigure}    

    \begin{subfigure}{0.33\linewidth}
        \centering
        \includegraphics[width=\linewidth]{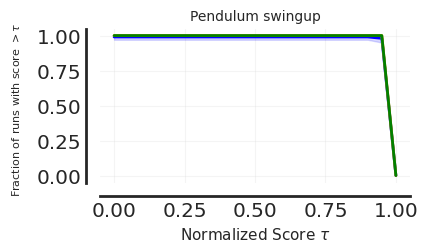}
    \end{subfigure}
    \hfill
    \begin{subfigure}{0.3\linewidth}
        \centering
        \includegraphics[width=\linewidth]{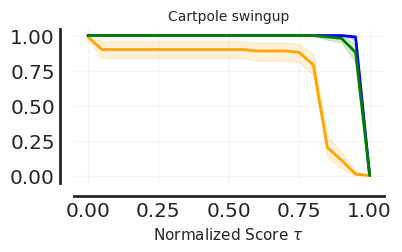}
    \end{subfigure}
    \hfill
    \begin{subfigure}{0.3\linewidth}
        \centering
        \includegraphics[width=\linewidth]{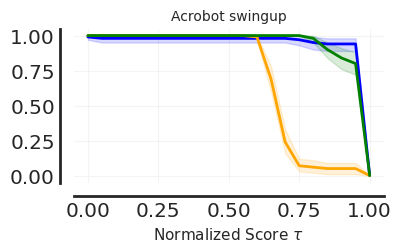}
    \end{subfigure}    
    \begin{subfigure}{\textwidth}
        \centering
        \includegraphics[width=0.6\linewidth]{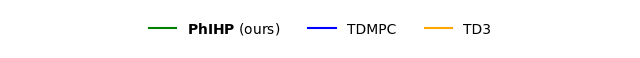}    
    \end{subfigure}
    \caption{Performance profiles of PhIHP and baselines on individual tasks (10 runs). Confidence intervals are estimated using the percentile bootstrap with stratified sampling \citep{agarwal2021deep}. PhIHP shows a better robustness to outliers.}
    \label{fig:rliable_pp_all}
\end{figure}

In this section, we compare performance metrics on individual classic control tasks. We estimate confidence intervals by using the percentile bootstrap with stratified sampling \citep{agarwal2021deep}.\\
We show in \cref{fig:rliable_all} a comparison of the median, interquartile median (IQM), mean performance, and optimality gap of PhIHP and baselines.  PhIHP matches or outperforms the performance of TD-MPC and TD3 in all tasks except in Cartpole swingup. PhIHP shown to be robust to outliers compared to TD-MPC with shorter confidence intervals.\\
Moreover, \cref{fig:rliable_pp_all} shows the performance profiles of PhIHP and baselines. PhIHP shows better robustness to outliers.\\ 

\newpage
\section{Hyperparameter sensitivity analysis}
\label{sec:hpsensib}
\begin{figure}[h]
    \rotatebox{90}{~~~~~ Pendulum}\hfill
    \begin{subfigure}{0.21\linewidth}
        \centering
        \includegraphics[width=\linewidth]{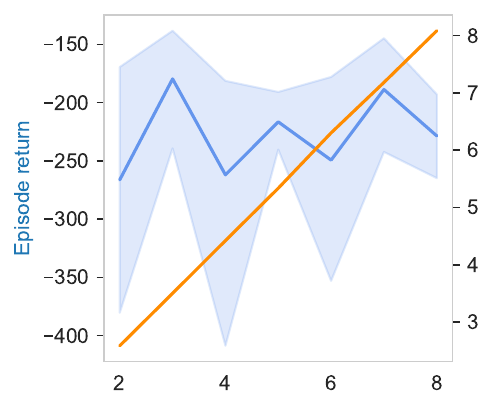}
    \end{subfigure}
    \hfill
    \begin{subfigure}{0.21\linewidth}
        \centering
        \includegraphics[width=\linewidth]{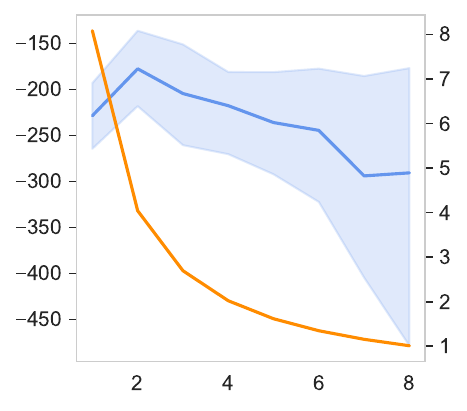}
    \end{subfigure}
    \hfill
    \begin{subfigure}{0.21\linewidth}
        \centering
        \includegraphics[width=\linewidth]{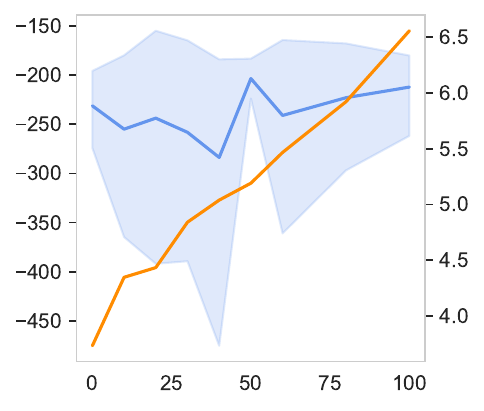}
    \end{subfigure}
    \hfill
    \begin{subfigure}{0.21\linewidth}
        \centering
        \includegraphics[width=\linewidth]{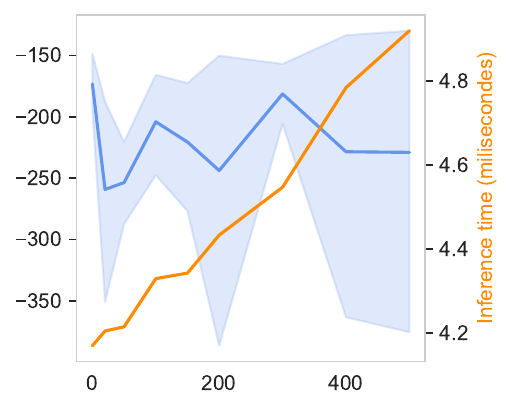}
    \end{subfigure}
    \vfill

    \rotatebox{90}{Pendulum swingup}\hfill
    \begin{subfigure}{0.21\linewidth}
        \centering
        \includegraphics[width=\linewidth]{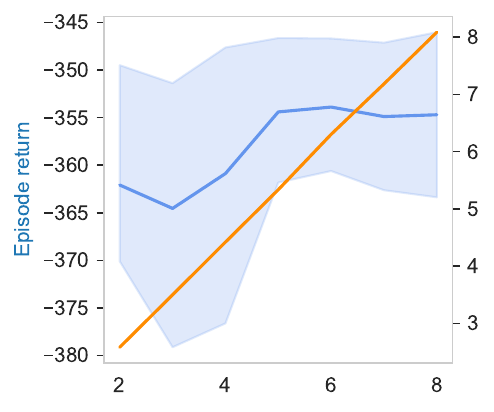}
    \end{subfigure}
    \hfill
    \begin{subfigure}{0.21\linewidth}
        \centering
        \includegraphics[width=\linewidth]{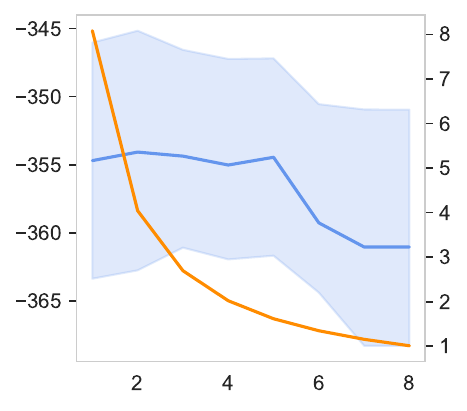}
    \end{subfigure}
    \hfill
    \begin{subfigure}{0.21\linewidth}
        \centering
        \includegraphics[width=\linewidth]{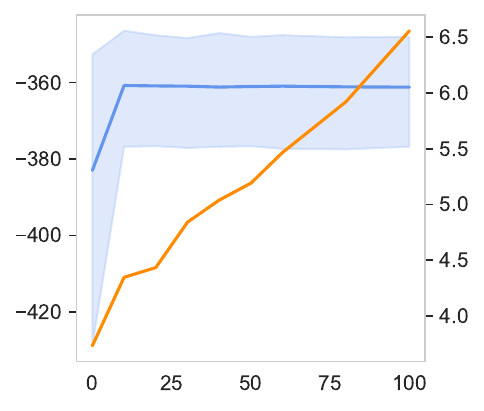}
    \end{subfigure}
    \hfill
    \begin{subfigure}{0.21\linewidth}
        \centering
        \includegraphics[width=\linewidth]{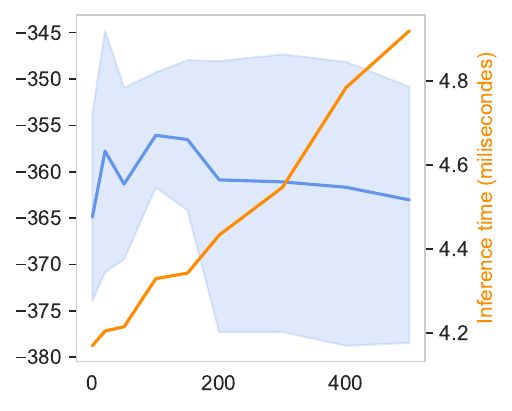}
    \end{subfigure}
    \vfill

    \rotatebox{90}{~~~~~~ Cartpole}\hfill
    \begin{subfigure}{0.21\linewidth}
        \centering
        \includegraphics[width=\linewidth]{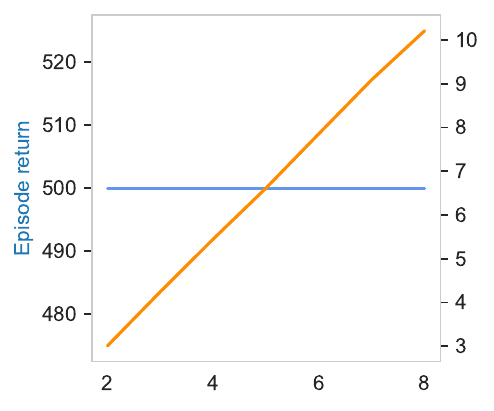}
    \end{subfigure}
    \hfill
    \begin{subfigure}{0.21\linewidth}
        \centering
        \includegraphics[width=\linewidth]{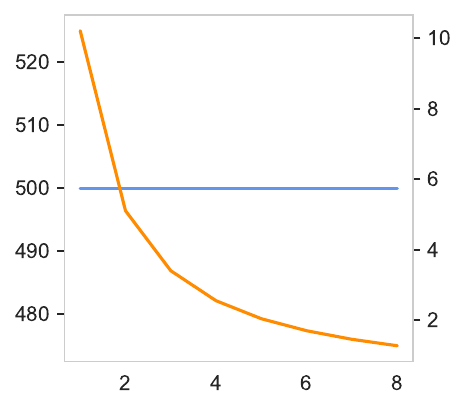}
    \end{subfigure}
    \hfill
    \begin{subfigure}{0.21\linewidth}
        \centering
        \includegraphics[width=\linewidth]{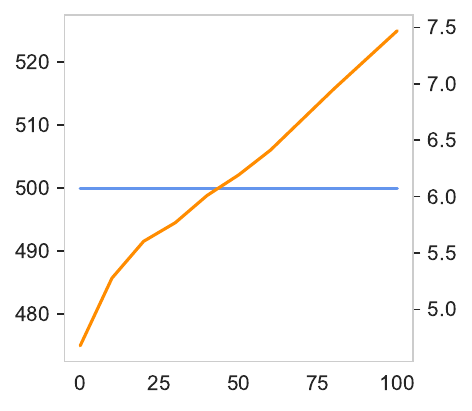}
    \end{subfigure}
    \hfill
    \begin{subfigure}{0.21\linewidth}
        \centering
        \includegraphics[width=\linewidth]{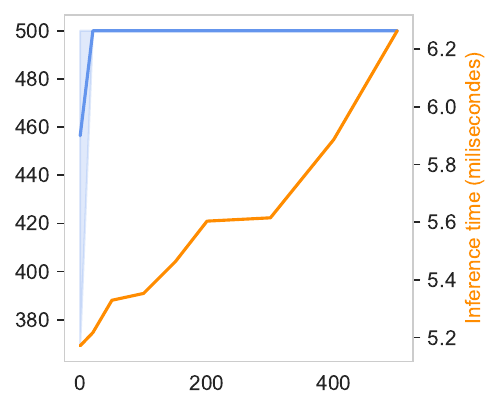}
    \end{subfigure}
    \vfill

    \rotatebox{90}{~Cartpole swingup}\hfill
    \begin{subfigure}{0.21\linewidth}
        \centering
        \includegraphics[width=\linewidth]{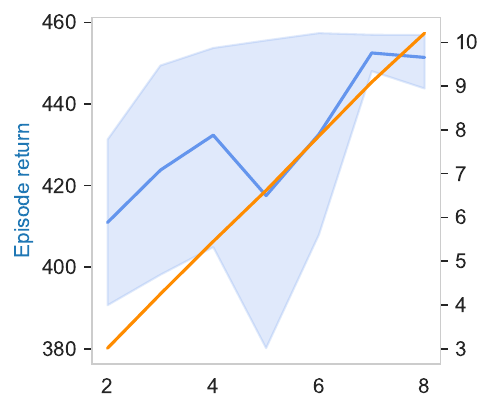}
    \end{subfigure}
    \hfill
    \begin{subfigure}{0.21\linewidth}
        \centering
        \includegraphics[width=\linewidth]{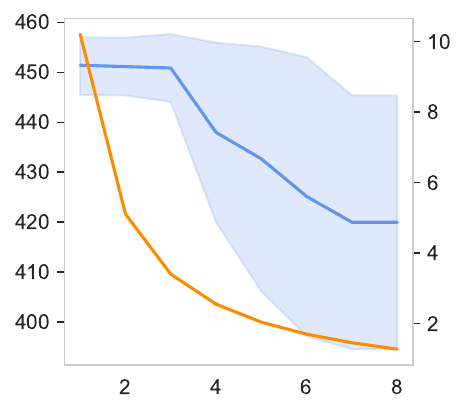}
    \end{subfigure}
    \hfill
    \begin{subfigure}{0.21\linewidth}
        \centering
        \includegraphics[width=\linewidth]{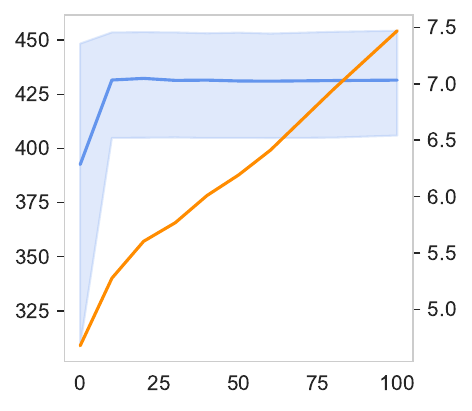}
    \end{subfigure}
    \hfill
    \begin{subfigure}{0.21\linewidth}
        \centering
        \includegraphics[width=\linewidth]{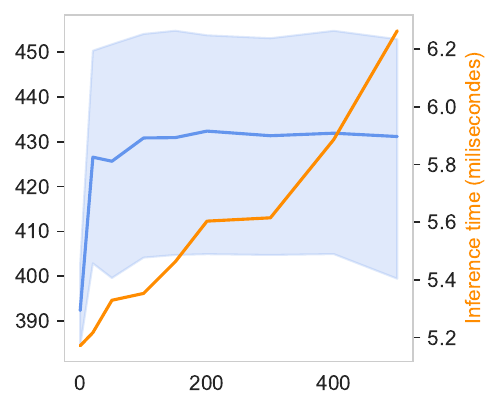}
    \end{subfigure}
    \vfill

    \rotatebox{90}{~ Acrobot swingup}\hfill
    \begin{subfigure}{0.21\linewidth}
        \centering
        \includegraphics[width=\linewidth]{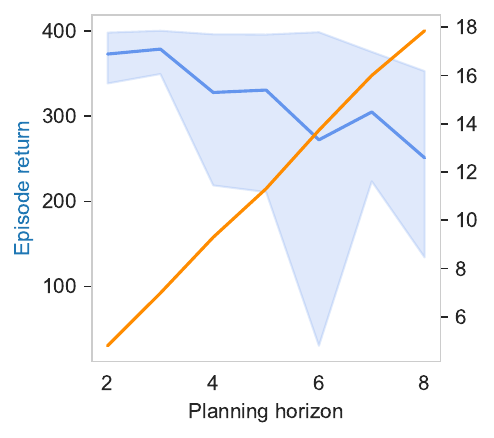}
    \end{subfigure}
    \hfill
    \begin{subfigure}{0.21\linewidth}
        \centering
        \includegraphics[width=\linewidth]{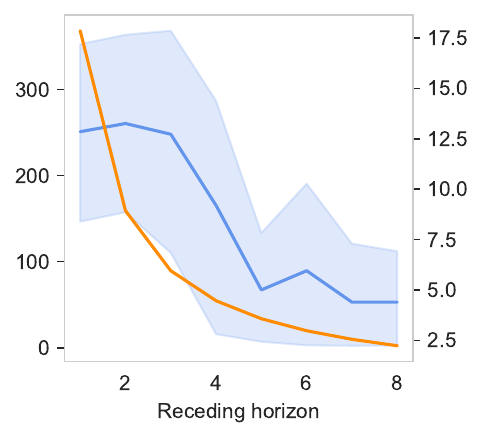}
    \end{subfigure}
    \hfill
    \begin{subfigure}{0.21\linewidth}
        \centering
        \includegraphics[width=\linewidth]{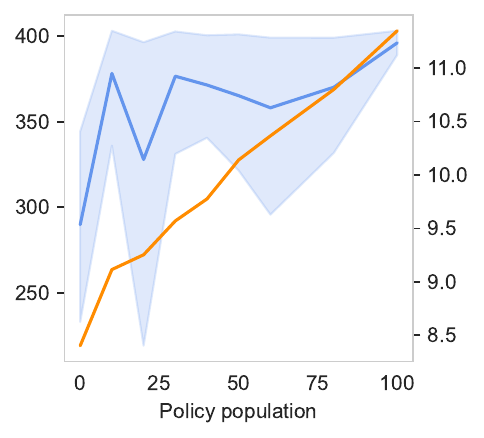}
    \end{subfigure}
    \hfill
    \begin{subfigure}{0.21\linewidth}
        \centering
        \includegraphics[width=\linewidth]{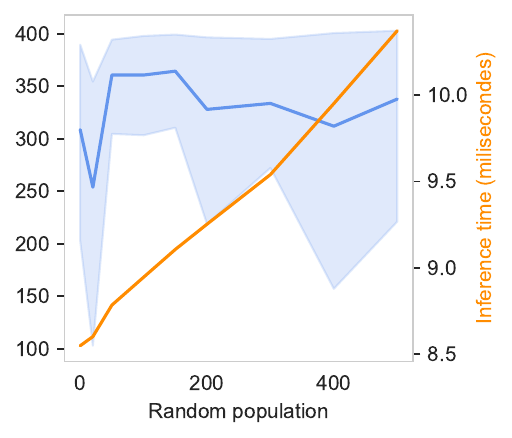}
    \end{subfigure}
    \begin{subfigure}{\textwidth}
        \centering
        \includegraphics[width=0.6\linewidth]{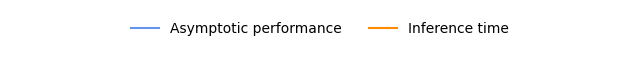}    
    \end{subfigure}
    \caption{Impact of varying planning hyperparameters on asymptotic performance and inference time on individual tasks.}
    \label{fig:analysiscomp}
\end{figure}

We investigate the impact of varying controller hyper-parameters on the performance and inference time of PhIHP.
We first study the impact of varying planning horizons and receding horizons (from 1 to 8). We note that planning over longer horizons generally leads to better performance, however, the performance slightly drops in Acrobot-swingup for planning horizon $H > 4$ (\cref{fig:analysiscomp}). We explain this by the compounding error effect on complex dynamics. Unsurprisingly, lower receding horizons always improve the performance because the agent benefits from replanning.\\
For the impact of the population size, \cref{fig:analysiscomp} shows that excluding the policy (policy-population = 0)  from planning degrades the performance, and increasing it under 10 does not have a significant impact. Moreover, excluding random actions (random-population = 0)  from planning degrades the performance.\\
Unsurprisingly, the inference time increases with an increase in both the planning horizon and the population size. Conversely, it decreases when the receding horizon increases.

\end{document}